\documentclass[11pt]{article}

\usepackage[preprint]{acl}

\usepackage{times}
\usepackage{latexsym}
\usepackage{makecell}
\usepackage{bm}
 \usepackage{amsfonts}
\usepackage[T1]{fontenc}
\usepackage{enumitem}

\usepackage[utf8]{inputenc}

\usepackage{microtype}

\usepackage{inconsolata}

\usepackage{xcolor}

\usepackage{hyperref}
\usepackage{amssymb} % For \checkmark
\usepackage{xcolor}  % For \textcolor
\usepackage{graphicx}
\usepackage{listings}
\usepackage{comment}
\usepackage[most]{tcolorbox}
\tcbset{sharp corners, boxrule=0.4pt, colback=gray!5, colframe=gray!60}
\usepackage[utf8]{inputenc}
\usepackage[vskip=0pt]{quoting}
\usepackage{stackengine}
\def\yenrule{\rule{1.3ex}{.1ex}}
\def\textyen{\renewcommand\stacktype{L}\stackon[.4ex]{\stackon[.65ex]{Y}{\yenrule}}{\yenrule}}
\title{GSM-SEM: Benchmark and Framework for Generating Semantically Variant Augmentations}

\author{
  \textbf{Jyotika Singh},
  \textbf{Fang Tu},
\textbf{Aziza Mirsaidova},
\textbf{Amit Agarwal},
\\\textbf{Hitesh Laxmichand Patel},
\textbf{Sandip Ghoshal},
  \textbf{Miguel Ballesteros},
  \textbf{Karan Dua},
  \\\textbf{Yassine Benajiba},
  \textbf{Weiyi Sun},
    \textbf{Tao Sheng},
  \textbf{Graham Horwood},
  \textbf{Sujith Ravi},
  \textbf{Dan Roth}
\\
  Oracle AI
\\
  \small{
    \textbf{Correspondence:} \href{mailto:jyotika.s.singh@oracle.com}{jyotika.s.singh@oracle.com}
  }
}

\begin{document}
\maketitle
\begin{abstract}
Benchmarks like GSM8K are popular measures of mathematical reasoning, but leaderboard gains can overstate true capability due to memorization of fixed test sets. Most robustness variants apply surface-level perturbations (paraphrases, renamings, number swaps, distractors) that largely preserve the underlying facts, and static releases can themselves become memorization targets over time.
We introduce \textbf{GSM-SEM}, a reusable and stochastic framework for generating \emph{\textbf{sem}antically diverse} benchmark variants with substantially higher semantic variance than prior approaches. GSM-SEM perturbs problem statements by modifying entities, attributes, and/or relationships, frequently altering underlying facts and requiring models to recompute solutions under new conditions, while constraining generation to preserve the original calculations/answer and approximate problem difficulty.
GSM-SEM generates fresh variants on each run without requiring re-annotation, reducing reliance on static public benchmarks for evaluation and thereby lowering the bias of memorization.
We apply GSM-SEM on GSM8K and two existing variation suites (GSM-Symbolic and GSM-Plus), producing \textbf{GSM8K-SEM}, \textbf{GSM-Symbolic-SEM}, and \textbf{GSM-Plus-SEM}. Evaluating 14 SOTA LLMs, we observe consistent performance drops with larger decline when semantic perturbations are coupled with symbolic/plus variations (average drop rate \textasciitilde28\% in maximum strictness configuration of GSM-SEM). We publicly release the three SEM variants as fully human-validated datasets. 
Finally, to demonstrate applicability beyond GSM-style math problems, we apply GSM-SEM to additional benchmarks including BigBenchHard, LogicBench, and NLR-BIRD.
\end{abstract}
\section{Introduction}

Benchmark leaderboards are widely used to track progress in language-model reasoning, with gains often interpreted as evidence of improved mathematical and logical competence. Yet a growing body of work suggests that high benchmark scores may reflect memorization, contamination, or exploitation of dataset-specific artifacts rather than robust generalization~\citep{sclarquantifying,gonen2023demystifyingprompts}. This concern is especially salient for widely used datasets such as GSM8K~\citep{cobbe2021trainingverifierssolvemath}, where repeated exposure through training mixtures and derivative resources can make static evaluation sets progressively less diagnostic.
\begin{figure}
    \includegraphics[width=1\linewidth]{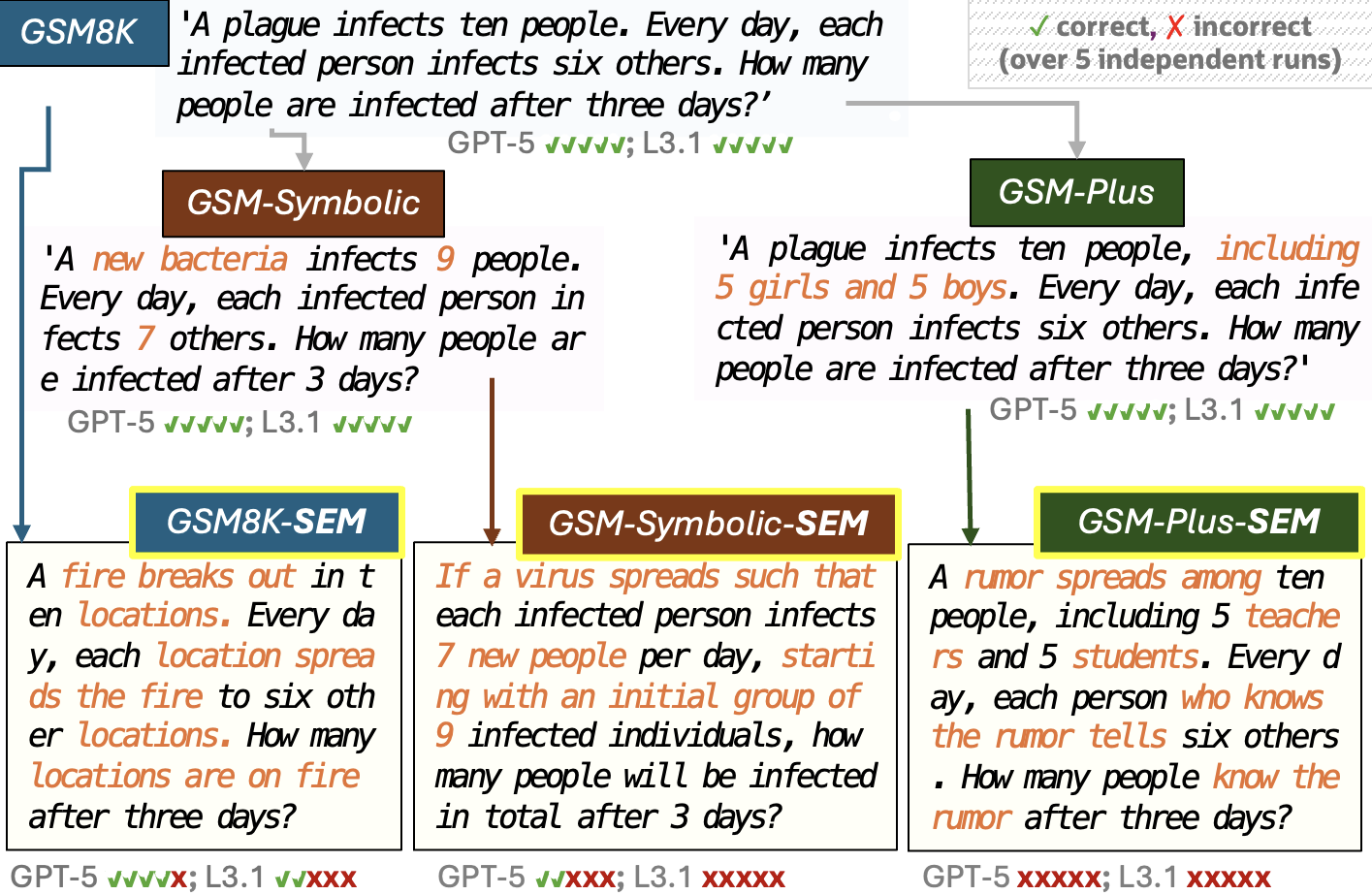}
    \caption{Example perturbations and per-run accuracy. Top: original GSM8K problem. Middle: GSM‑Symbolic and GSM‑Plus rewrites. Bottom: GSM-SEM variants (orange highlights indicate edited spans). For each panel, accuracy across five independent runs is shown (\textcolor{green}{\checkmark} correct, \textcolor{red}{x} incorrect), illustrating higher failures on SEM variants. L3.1 = Llama‑3.1‑405B-Ins; GPT‑5 uses medium/default reasoning effort.}
    \label{fig:examples}
\end{figure}

A common response has been to introduce robustness variants via paraphrasing, entity renaming, and numerical substitutions~\citep{mirzadehgsm, li2024gsm,lunardi2025robustnessreliabilitybenchmarkbasedevaluation,wang2024rupbenchbenchmarkingreasoningperturbations,rabinovich2023predicting, rajmeasuring, sclarquantifying}. While these perturbations are useful for testing sensitivity to wording and superficial cues, they typically preserve the original factual backbone and relational structure of each problem. As a result, they often probe invariance to surface form more than adaptability to meaningfully different semantic conditions. %Moreover, once such variants are released, they can themselves become static memorization targets, again weakening their value for long-term evaluation.
Moreover, most evaluations focus on older or smaller models, where effects are large, while the few newer SOTA models of the time assessed show smaller declines, leaving open whether recent SOTA systems have developed more robust reasoning capabilities.

We address this gap with \textbf{GSM-SEM}, a reusable framework for generating \textbf{semantically diverse} variants with substantially higher semantic variance than prior transformations (Figure~\ref{fig:examples}), frequently altering underlying facts requiring models to recompute solutions under new conditions, while constraining generation to preserve the original ground-truth final answer (and associated computation). % and approximate problem difficulty.
Importantly, GSM-SEM is \emph{stochastic and reusable}: it generates fresh variants on each run without re-annotation (due to ground-truth answer and calculation preservation), reducing reliance on static public benchmarks and thereby lowering evaluation bias from memorization as models and training corpora evolve. 
Our main contributions are:
\begin{itemize}[leftmargin=*]
\item \textbf{GSM-SEM}: A stochastic, reusable framework for generating semantically diverse benchmark variants %by perturbing entities, attributes, and relationships 
while preserving the original calculations/answer and targeting similar difficulty.
\item \textbf{Human-validated benchmark release}: Public release of three fully human-validated benchmarks - GSM8K-SEM, GSM-Symbolic-SEM, GSM-Plus-SEM. \footnote{\url{https://sites.google.com/view/gsm-sem/home}}
\item \textbf{Comprehensive evaluation}: Study of 14 SOTA LLMs across model families, showing consistent performance degradation in GSM-SEM variants relative to the original benchmarks, with amplified failures when semantic perturbations are coupled with symbolic and plus variations.
\item \textbf{Extension}: Demonstration of the framework’s applicability to out-of-domain non-GSM samples from BigBenchHard~\cite{suzgun2022challenging}, LogicBench~\cite{parmar-etal-2024-logicbench}, and NLR-BIRD~\cite{singh-etal-2025-llms}.
\end{itemize}

\section{Related Work}

GSM8K is a widely used benchmark for evaluating LLMs on grade-school mathematical reasoning. However, its extensive adoption and relatively standardized problem format have raised concerns that leaderboard gains may partially reflect overfitting, contamination, or memorization rather than robust reasoning~\citep{xie2025memorization,zhang2024careful}. A large body of work shows that small changes to problem statements - including entity substitutions, paraphrasing, and the insertion of irrelevant context - can induce substantial performance drops~\citep{shi2023large,li2024gsm,wu-etal-2024-reasoning,zhou-etal-2024-paraphrase,srivatsa-kochmar-2024-makes,bhuiya-etal-2024-seemingly,shi-freda-2023-irrelevant-context,karim2025lostculturaltranslationllms}, suggesting that strong GSM8K accuracy alone can be a brittle indicator of generalization.

\textbf{Adversarial context and distractors:}
One line of work probes fragility by introducing adversarial or irrelevant information. GSM-IC (Irrelevant Context)~\citep{shi2023large} demonstrates that LLMs may incorporate distractor sentences into intermediate reasoning, degrading performance. iGSM~\citep{ye2024physics} further studies sensitivity to changes in prompting and problem structure. While these are valuable for isolating specific failure modes, they often target a narrower set of perturbations relative to more recent, multi-phenomena suites.

\textbf{Comprehensive perturbation suites:}
GSM-Plus~\citep{li2024gsm} expands GSM8K with a large collection of derived variants spanning multiple perturbation categories (e.g., rephrasing, distractors, numerical variation, arithmetic/operation changes, and higher-order reasoning demands). This breadth makes GSM-Plus a useful modern testbed that covers distractor-style failures (including those emphasized by GSM-IC) while also probing additional robustness dimensions. Many GSM-Plus transformations modify the underlying computation and therefore need new ground-truth answers.

\textbf{Symbolic synthesis:}
A complementary line addresses the instability of single-point benchmarks by generating \emph{distributions} of variants. GSM-Symbolic~\citep{mirzadehgsm} uses symbolic templates to synthesize controlled instantiations of problems by varying names and numbers, enabling statistical evaluation across many realizations. This template-based view reveals that model accuracy can be highly sensitive to seemingly minor changes (e.g., specific values and therefore calculations) and can support more rigorous measurement than small-scale iterative perturbations. Because GSM-Symbolic typically changes numbers, variants need updated ground-truth answers.

\begin{figure*}[!t]
    \centering
    \includegraphics[width=1.0\linewidth]{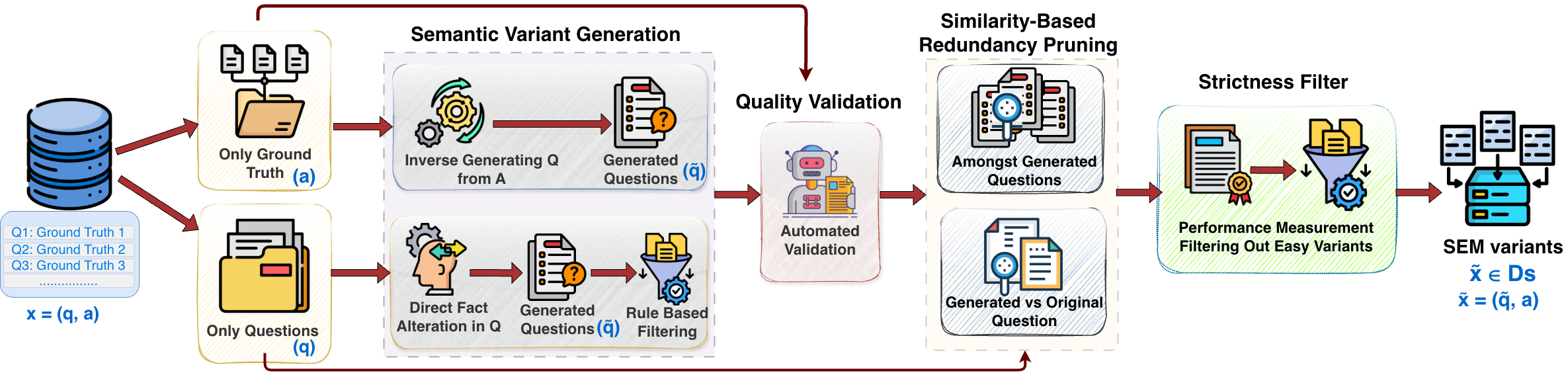}
    \caption{GSM-SEM: \textbf{Sem}antic variant generation pipeline. The pipeline comprises semantic variant generation, automated quality validation steps, similarity-based redundancy pruning, and strictness filtering.}
    \label{fig:gsmrepipeline}
\end{figure*}

\textbf{GSM-SEM} targets a different axis than prior variants: \emph{high semantic variance} via edits to the narrative, while constraining generation to preserve the original answer and approximate difficulty. This addresses two practical limitations of many existing benchmarks: (i) released variants are static and can become memorization targets over time; and (ii) extending them with fresh, comparable perturbations is often infeasible or requires re-annotating ground-truth answers, which is hard to do reliably at scale in an automated fashion. By generating fresh variants while keeping the final answer fixed, GSM-SEM supports re-runnable evaluation without re-annotation and enables \emph{compositional} robustness testing: it can be applied directly to GSM8K and layered on top of existing robustness suites. We focus in particular on GSM-Plus and GSM-Symbolic because they provide complementary substrates for such composition---a broad, multi-phenomena perturbation suite (GSM-Plus) and a template-driven framework enabling evaluation across different calculations for the same problem (GSM-Symbolic). Extended related work / methodological details are shared in Appendix~\ref{sec:relatedwork}.

\section{Methods - GSM-SEM pipeline}

The GSM-SEM framework (Figure~\ref{fig:gsmrepipeline}) generates semantic variations of GSM-based datasets (GSM8k, GSM-Symbolic, GSM-Plus) through a dual curation and validation process followed by pruning the generated set based on similarity-based redundancy and tunable strictness filtering. These perturbations maintain the same answers as the original data, without altering the calculation and values required for the solutions. 
This method is further applied to out-of-domain datasets (BigBenchHard, LogicBench, NLR-BIRD) to demonstrate potential for broader applicability beyond GSM. Appendix~\ref{sec:datasettings} contains details of prompts, models, and settings.

\paragraph{Semantic Variant Generation}
Let $x=(q,a)$ denote an original sample with question $q$ and ground-truth numeric answer $a$. We generate semantically diverse variants $\tilde{x}=(\tilde{q},a)$ using two complementary augmentation strategies. The two strategies are designed to be complementary. One conditions on the solution and answer, anchoring the computation while enabling larger semantic variation. Other conditions on the original question, preserving fine-grained structural constraints while modifying narrative elements. These streams cover different points on the diversity–control tradeoff, and combining them reduces failure modes of either alone. First, given $(q,a)$, we prompt diverse LLMs to inverse generate a new question $\tilde{q}$ (`reverse engineering' questions from answers) that preserves the underlying numerical computation implied by $a$ while allowing the surface form and contextual facts to change substantially (e.g., from `new bacteria infects' to `virus spreads'; Figure~\ref{fig:examples}). As evaluation process compares only the final numeric answer (or also calculations for LLM-judge-based evaluation), the ground-truth value $a$ remains valid for these variants. Second, we prompt the LLM to keep all numerical values fixed while altering the scenario/theme in the question, thereby preserving the required computation but increasing semantic diversity (e.g., from `plague infects' to `rumor spreads' and `fire breaks out'; Figure~\ref{fig:examples}). We then apply a rule-based verifier that checks that all numeric values appearing in each $\tilde{q}$ match those in the corresponding original question $q$, to remove obvious numeric drift failures.

\paragraph{Automated Quality Validation}

We designed the automated validator by first having human reviewers assess 70 randomly sampled augmented variants, flagging both acceptable and flawed examples (details in Appendix~\ref{sec:dataqa}). Two independent reviewers labeled the data for quality and logical coherence, and these annotations (on samples distinct from those used in our main evaluation) were used as input to engineer a judge LLM prompt to mirror the same decision criteria for tuning the pipeline.

\paragraph{Similarity-Based Redundancy Pruning}
Let $x$ denote a base sample and $\tilde{x}$ a candidate variant. We compute the cosine-based similarity $s(x,\tilde{x})$ based on term frequency vectors and discard candidates with $s(x,\tilde{x})>0.85$, retaining only variants that are sufficiently divergent from the base sample. To further promote diversity among retained variants, we additionally perform set-level de-duplication. Specifically, given the current retained set $\mathcal{S}$, we accept a candidate $\tilde{x}$ only if $s(\tilde{x}, x') \le 0.85$ for all $x' \in \mathcal{S}$. This criterion ensures that each added variant is distinct both from the original sample and from previously retained variants.

\paragraph{Strictness Filter}
\label{sec:sf}
This step applies a strictness filter to control how challenging the final samples should be to further filter easy and redundant variants.
We first evaluate $\tilde{x}$ using a held-out set of reference models $\mathcal{M}_{\text{hold}}$ and compute a cross-model consistency score $c(\tilde{x})$ that captures agreement in correctness across $\mathcal{M}_{\text{hold}}$. Variants with unanimous agreement are treated as potentially \emph{easy}, as they may either (i) oversimplify the underlying task and thus fail to probe robustness, or (ii) remain insufficiently perturbed relative to $x$. To mitigate both cases, for high-consistency variants we apply a stricter semantic pruning rule by constraining the semantic similarity $s(x,\tilde{x})$ to lie within an acceptance window $[\alpha,\beta]$; variants with $s(x,\tilde{x}) \notin [\alpha,\beta]$ are discarded. The held-out model set is $\mathcal{M}_{\text{hold}}=\{\text{Llama-3.3-70B}, \text{GPT-4o}, \text{GPT-4o-mini}\}$. These models are used exclusively for filtering and are excluded from the main evaluation in Section~5, which is conducted on a disjoint set of 14 models.

We summarize the validation and filtering pipeline as the following acceptance criterion for adding a candidate variant $\tilde{x}$ to the retained set $\mathcal{D}_s$,
\begin{equation*}\label{eq:accept}
\tilde{x}\in\mathcal{D}_s\ \leftrightarrow\
\begin{aligned}[t]
& s(x,\tilde{x}) \le \tau \\
&\wedge\ \max_{x'\in\mathcal{S}} s(\tilde{x},x') \le \tau \\
&\wedge\ \Big(cm(\tilde{x})<1\ \vee\ s(x,\tilde{x})\in[\alpha,\beta]\Big)
\end{aligned}
\end{equation*}
\noindent
where\
$s(\cdot,\cdot)$ is cosine similarity between variants and original question, and among different variants;
$cm(\tilde{x})\in[0,1]$ is the cross-model consistency score computed on a held-out
model set $\mathcal{M}_{\text{hold}}$.
We use $\tau=0.85$ and present results across several configurations of $[\alpha,\beta]$ for filter strictness. Section~\ref{sec:resultsmain} shows impact of different settings and we adopt a moderate $[\alpha,\beta]$ 0.4–0.6 (rather than removing all unanimously-correct variants) for detailed results shared in the main paper, with more results in the appendix for other configurations. %, but the pipeline can be re-run with stricter thresholds as needed.
Figure~\ref{fig:sffiltervalues} in Appendix~\ref{sec:sfsettings} details all \emph{Strictness Filter} settings.
\section{Datasets}

\subsection{GSM8K-SEM, GSM-Symbolic-SEM,  GSM-Plus-SEM}
Using the GSM-SEM framework, we generated the GSM8K-SEM, GSM-Symbolic-SEM, and GSM-Plus-SEM datasets, which comprise multiple perturbed samples per original base data sample ~\ref{sec:gsm8k_sample}, resulting in a total of 685, 2525, and 1436 samples, respectively. Dataset size details and breakdown by \emph{Strictness Filter} are shared in Appendix~\ref{sec:datasetsize}.

\paragraph{Fully Human Validated Datasets}

We subjected these full datasets generated by the GSM-SEM pipeline to human validation at the sample level to ensure that all reported results are based exclusively on a vetted subset. Each sample was independently reviewed by two annotators for overall quality and logical coherence, drawing from a pool of six reviewers in total (see Appendix~\ref{sec:dataqa} for full statistics and details). Fewer than 3\% of samples were rated below the high-quality threshold, further supporting the robustness of our generation approach. All published datasets and results include only samples that have been validated as good by humans.

\begin{figure}[!t]
    \centering
    \includegraphics[width=1.0\linewidth]{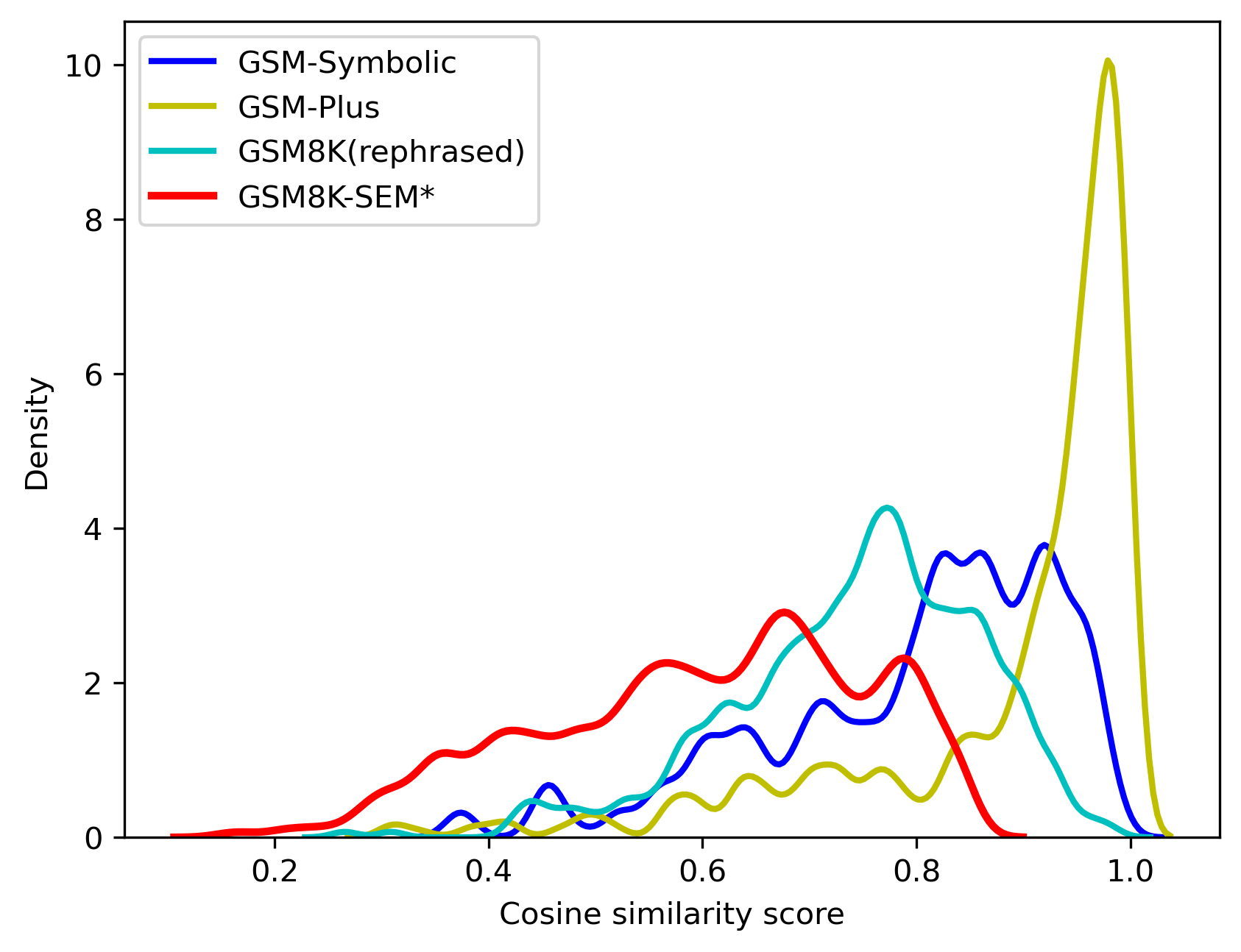}
    \caption{Cosine similarity distribution of GSM8K variants with respect to GSM8K, where GSM8K-SEM shows higher semantic divergence than other variants.}
    \label{fig:cossimtrend}
\end{figure}

\paragraph{Semantic Similarity Distribution} As our goal is to create semantically different samples from the original dataset, Figure~\ref{fig:cossimtrend} shows the count-based cosine similarity distribution between the GSM8K-SEM and the GSM8K dataset. 
For comparison, we also report the similarity distribution for a paraphrased version of GSM8K. GSM8K-SEM is shifted toward lower similarity, indicating that it differs more substantially from the original dataset and thus is not equivalent to paraphrasing. We also show the count-based cosine similarity distribution between GSM-Symbolic and GSM8K, and between GSM-Plus and GSM8K. As GSM-Symbolic essentially only swaps entities from the original dataset, its similarity tends to be higher than paraphrased queries. GSM-Plus shows high similarity, consistent with many variants introducing only minimal edits. 
Figure~\ref{fig:cossimtrend} reports GSM8K-SEM with strictness filter $[\alpha,\beta]~0,1$. Figure~\ref{fig:cossimSFall} shows how the similarity distributions change under different strictness filter settings in Appendix~\ref{sec:cosine}.

For a more fine-grained comparison, we also compare cosine similarity using sentence embeddings (all-MiniLM-L6-v2) (see Figure~\ref{fig:cossimbert}, Appendix~\ref{sec:cosine}). Paraphrased questions have the highest semantic similarity to the originals, as they convey the same content. GSM-Symbolic-SEM, which incorporates entity swaps and semantic modifications, exhibits reduced similarity and SEM variants continue to show lowest similarity. 

\subsubsection{Evaluation set-up and metrics}
We measure accuracy by running each model at least five times per sample ($R\ge5$), averaging correctness across runs, and then averaging over all samples to obtain the final accuracy score. 
Let $\mathcal{D}_s$ be the set of SEM variants, where each element is a pair $(\tilde{x},y)$.
\[\mathrm{Acc}
=\frac{1}{|\mathcal{D}_s|}\sum_{(\tilde{x},y)\in\mathcal{D}_s}
\left(\frac{1}{R}\sum_{r=1}^{R}\mathbb{I}\!\left[\mathrm{LM}_r(\tilde{x})=y\right]\right)\]
where $\mathbb{I}[\cdot]$ is the indicator function and $\mathrm{LM}_r(\tilde{x})$ denotes the model output on run $r$. Acc $\Delta$ is the variant accuracy subtracted by original accuracy.

Correctness is determined using a hybrid evaluator: rule-based numeric extractor complemented by LLM-as-a-judge. We found that rule-based evaluators used in prior work at times fail to extract valid answers (yield false negatives), whereas the LLM judge substantially improves coverage and reliability \cite{yuxfinder}. See Appendix~\ref{sec:datasettings} for prompt/settings. 

We also measure the performance drop rate (PDR) metric to show the relative performance decline on
variants compared to the baseline samples.
\[
\mathrm{PDR}
= 1 -
\frac{\frac{1}{|\mathcal{D}_s|}\sum_{(\tilde{x},y)\in \mathcal{D}_a} \mathbb{I}\!\left[\mathrm{LM}(\tilde{x})=y\right]}
     {\frac{1}{|\mathcal{D}|}\sum_{(x,y)\in \mathcal{D}} \mathbb{I}\!\left[\mathrm{LM}(x)=y\right]}
\]
where $\mathcal{D}_s$ and $\mathcal{D}$ represent the SEM variants and baseline datasets, respectively.

We report these metrics on original GSM8K dataset, GSM-Symbolic, GSM-Plus and their SEM variants, all evaluated under zero-shot inference. Additional experiments using 8-shot Chain-of-prompting yielded trends consistent with the zero-shot setting, so we focus on zero-shot results here (details in Appendix~\ref{sec:cot}).

\subsection{Out-of-domain datasets}
To assess broader applicability of GSM-SEM pipeline on non-GSM data, we evaluated the SEM pipeline by applying it to several out of domain datasets, each representing distinct type of data and reasoning difficulty. These included BigBench-Hard~\cite{suzgun2022challenging, srivastava2023beyond} (temporal sequences), LogicBench~\cite{parmar-etal-2024-logicbench} (first-order logic), and NLR-BIRD~\cite{singh-etal-2025-llms}. Each dataset diverges from the math-oriented GSM benchmark: 
BigBench-Hard presents lengthy sequences of temporal events, demanding more complex reasoning to identify the correct response; LogicBench entails multiple-choice questions testing first-order logic skills, such as hypothetical and disjunctive syllogisms, as well as various logical dilemmas; NLR-BIRD, a newer dataset, focuses on nuanced linguistic reasoning to present tables into natural language representations (NLR). For each dataset, example SEM variants generated are provided along with experimental details in Appendix~\ref{sec:apxood}.

\section{Results and Discussion}
\label{sec:resultsmain}

\begin{figure*}[!t]
    \centering
    \includegraphics[width=1\linewidth]{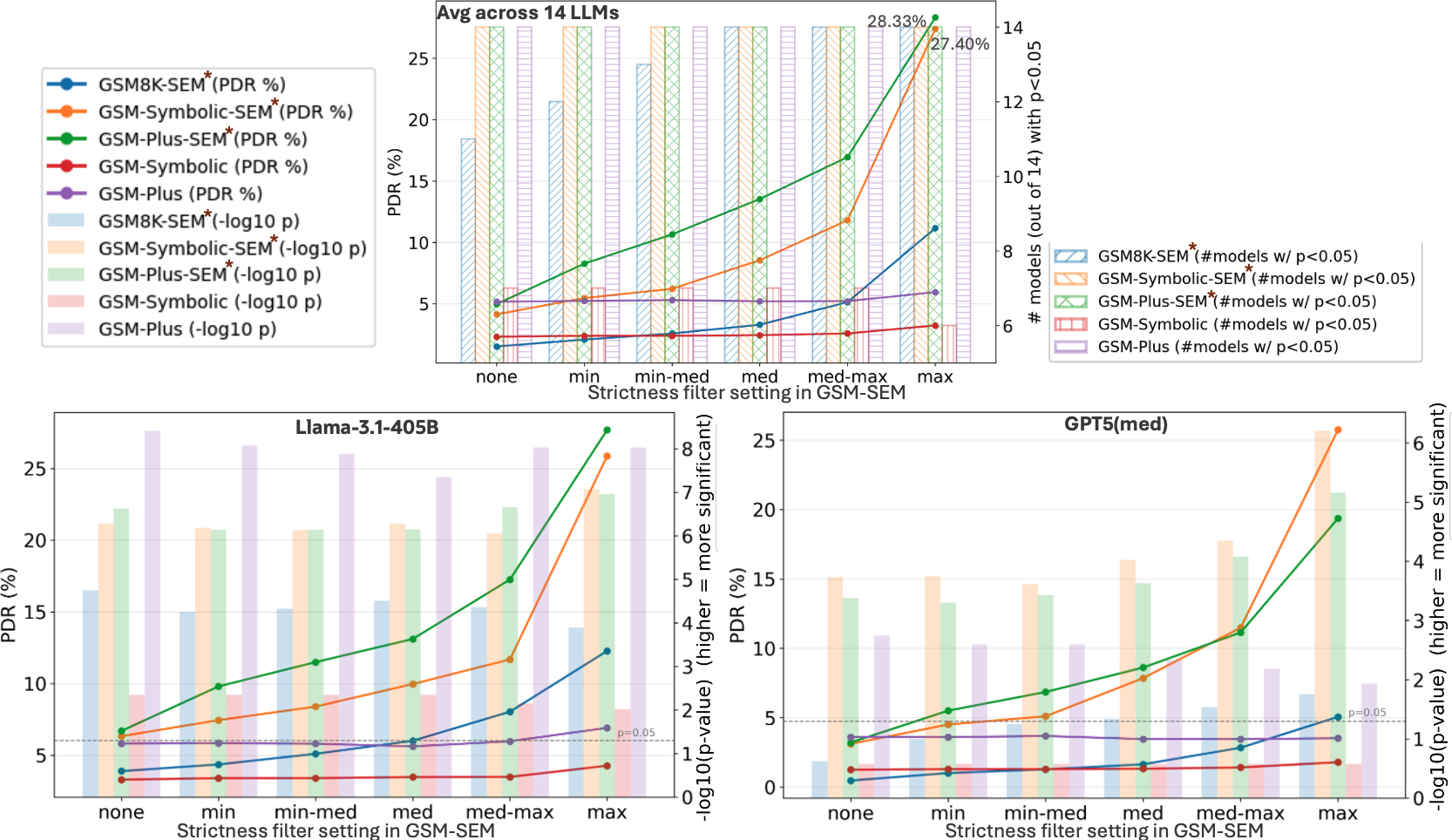}%resthrspdr.png}
    \caption{Effect of \emph{Strictness Filter} (Section~\ref{sec:sf}) on PDR\% (relative to GSM8K) and statistical significance. Filter settings: none (all samples kept; $[\alpha,\beta]$ 0-1), min ($[\alpha,\beta]$ 0.30–0.70), min-med (0.35–0.65), med (0.40–0.60), med-max (0.45–0.55), and max (all such samples filtered out). Variant dataset sizes across filters are shared in Table~\ref{tab:variant-strictness-sizes}.}
    \label{fig:respipelinethrs}
\end{figure*}

Figure~\ref{fig:respipelinethrs} shows the performance drop rate for different tightness of the GSM-SEM pipeline \emph{Strictness filter} step, at varying $[\alpha,\beta]$.
Table~\ref{tab:gsmresdetailed} presents the performance across 14 LLM settings on GSM8K, Symbolic, Plus, and their SEM variants for the \emph{med} configuration of the \emph{Strictness filter}.

\begin{table*}[!t]
\small
\centering
\begin{tabular}{p{2.0cm}cc|cc|cc}
  & Paraphr & GSM8K-\textbf{SEM} & GSM-Symb & Symb-\textbf{SEM} & GSM-Plus & Plus-\textbf{SEM} \\
  & Acc$\Delta$ [PDR] &  Acc$\Delta$ [PDR] &  Acc$\Delta$ [PDR] &  Acc$\Delta$ [PDR] &  Acc$\Delta$ [PDR] &  Acc$\Delta$ [PDR] \\
\hline
Grok3           & 1.04 [-1.58]  & -3.70 [2.89]{\tiny \textcolor{red}{$\bm{\downarrow}$}} & -0.76 [0.80] & -7.11 [7.52]{\tiny \textcolor{red}{$\bm{\downarrow}$}} & -3.77 [3.74] & -13.00 [13.46]{\tiny \textcolor{red}{$\bm{\downarrow}$}}  \\
Llama3.1-405    & -1.07 [0.62]  & -6.84 [6.01]{\tiny \textcolor{red}{$\bm{\downarrow}$}} & -3.40 [3.48] & -9.73 [9.97]{\tiny \textcolor{red}{$\bm{\downarrow}$}} & -5.81 [5.62] & -13.06 [13.11]{\tiny \textcolor{red}{$\bm{\downarrow}$}}  \\
Llama4-Mav      & -2.22 [1.79]  & -3.79 [2.85]{\tiny \textcolor{red}{$\bm{\downarrow}$}} & -3.48 [3.55] & -7.34 [7.50]{\tiny \textcolor{red}{$\bm{\downarrow}$}} & -6.42 [6.37] & -16.34 [16.47]{\tiny \textcolor{red}{$\bm{\downarrow}$}}  \\
Llama4-Scout    & -0.51 [0.57]  & -3.88 [4.10]{\tiny \textcolor{red}{$\bm{\downarrow}$}} & -2.76 [2.86] & -7.97 [8.26]{\tiny \textcolor{red}{$\bm{\downarrow}$}} & -6.93 [7.03] & -15.33 [15.81]{\tiny \textcolor{red}{$\bm{\downarrow}$}}  \\
GPT4.1          & -0.95 [0.51]   & -3.85 [2.98]{\tiny \textcolor{red}{$\bm{\downarrow}$}} & -3.33 [3.47] & -7.98 [8.31]{\tiny \textcolor{red}{$\bm{\downarrow}$}} & -4.47 [4.26] & -11.86 [11.87]{\tiny \textcolor{red}{$\bm{\downarrow}$}}  \\
GPT4.1-mini     & -1.77 [0.84]  & -2.68 [1.73]{\tiny \textcolor{red}{$\bm{\downarrow}$}} & -2.42 [2.51] & -6.83 [7.06]{\tiny \textcolor{red}{$\bm{\downarrow}$}} & -4.25 [4.05] & -12.19 [12.35]{\tiny \textcolor{red}{$\bm{\downarrow}$}}  \\
Gemini2.5-f     & -2.51 [2.12]  & -5.07 [4.23]{\tiny \textcolor{red}{$\bm{\downarrow}$}} & -3.11 [3.17] & -6.94 [7.16]{\tiny \textcolor{red}{$\bm{\downarrow}$}} & -6.66 [6.57] & -13.38 [13.34]{\tiny \textcolor{red}{$\bm{\downarrow}$}}  \\
Gemini2.5-f-l   & -1.95 [1.04]  & -4.44 [3.64]{\tiny \textcolor{red}{$\bm{\downarrow}$}} & -0.52 [0.55] & -7.27 [7.62]{\tiny \textcolor{red}{$\bm{\downarrow}$}} & -6.40 [6.27] & -14.05 [14.53]{\tiny \textcolor{red}{$\bm{\downarrow}$}}  \\
Gemini2.5-pro   & -0.82 [0.36]  & -3.49 [3.39]{\tiny \textcolor{red}{$\bm{\downarrow}$}} & -0.86 [0.88] & -6.66 [6.86]{\tiny \textcolor{red}{$\bm{\downarrow}$}} & -4.15 [4.07] & -11.53 [11.35]{\tiny \textcolor{red}{$\bm{\downarrow}$}}  \\
O3              & -0.19 [-0.29]  & -3.36 [2.68]{\tiny \textcolor{red}{$\bm{\downarrow}$}} & -2.18 [2.23] & -8.71 [8.91]{\tiny \textcolor{red}{$\bm{\downarrow}$}} & -5.06 [4.91] & -9.77 [9.45]{\tiny \textcolor{red}{$\bm{\downarrow}$}}  \\
GPT5(mnml)      & -1.16 [0.71]  & -3.37 [2.42]{\tiny \textcolor{red}{$\bm{\downarrow}$}} & -1.69 [1.73] & -8.68 [8.89]{\tiny \textcolor{red}{$\bm{\downarrow}$}} & -4.84 [4.46] & -12.86 [12.43]{\tiny \textcolor{red}{$\bm{\downarrow}$}}  \\
GPT5.1(mnml)    & -1.99 [1.58]  & -4.70 [3.83]{\tiny \textcolor{red}{$\bm{\downarrow}$}} & -2.15 [2.21] & -11.26 [11.61]{\tiny \textcolor{red}{$\bm{\downarrow}$}} & -6.02 [5.73] & -18.52 [18.69]{\tiny \textcolor{red}{$\bm{\downarrow}$}}  \\
GPT5(med)       & -0.07 [-0.42]  & -2.59 [1.64]{\tiny \textcolor{red}{$\bm{\downarrow}$}} & -1.28 [1.32] & -7.60 [7.85]{\tiny \textcolor{red}{$\bm{\downarrow}$}} & -3.83 [3.46] & -9.09 [8.61]{\tiny \textcolor{red}{$\bm{\downarrow}$}}  \\
GPT5.1(med)     & -2.26 [1.84]  & -4.47 [3.56]{\tiny \textcolor{red}{$\bm{\downarrow}$}} & -5.20 [5.32] & -11.80 [12.08]{\tiny \textcolor{red}{$\bm{\downarrow}$}} & -6.50 [6.18] & -18.40 [17.93]{\tiny \textcolor{red}{$\bm{\downarrow}$}}  \\
\hline
\hline
Avg Acc Drop
      & -1.17\% & \textbf{-4.02\%} & -2.36\% & \textbf{-8.28\%} & -5.37\% & \textbf{-13.53\%} \\
\hline
Avg PDR &
      0.69\% & \textbf{3.28\%} & 2.43\% & \textbf{8.54\%} & 5.19\% & \textbf{13.53\%} \\
\end{tabular}
\caption{Accuracy drop (in \%) and PDR compared to GSM8K across GSM-based datasets and our SEM variations. Positive PDR indicates performance decline. These results are on 100\% human-validated data samples. The bracket next to GPT5x reflects reasoning parameter setting (mnml=minimal; med=medium).}
\label{tab:gsmresdetailed}
\end{table*}

\paragraph{Performance drops in SEM variants} As can be seen in Table \ref{tab:gsmresdetailed}, there is a consistent decline in performance in each SEM variant compared to original baseline data. We performed a Wilcoxon test to determine whether SEM variants perform worse over the baselines with \textbf{statistical significance}. Our results show that 14/14 models demonstrate worse performance than baseline with statistical significance (p<0.05) for all SEM variants in the \emph{med} configuration (shown in Figure~\ref{fig:respipelinethrs}). Detailed statistics breakdown per model is shared in Appendix~\ref{sec:statdetailed}.

\paragraph{Impact of Compounded Perturbations}Figure~\ref{fig:respipelinethrs} and Table~\ref{tab:gsmresdetailed} show that GSM-Plus-SEM and GSM-Symbolic-SEM yield the highest accuracy drops across models among all variants, indicating that compounding semantic and other perturbations (symbolic and others in plus) amplify the challenge. Notably, GSM8K-SEM, GSM-Symbolic, and GSM-Plus alone also cause measurable accuracy reductions relative to the original GSM benchmark, but the semantic nature and effects of these perturbations differ considerably thus leading to compounded decline when combined.

\begin{figure}[!t]
    \centering
    \includegraphics[width=1\linewidth]{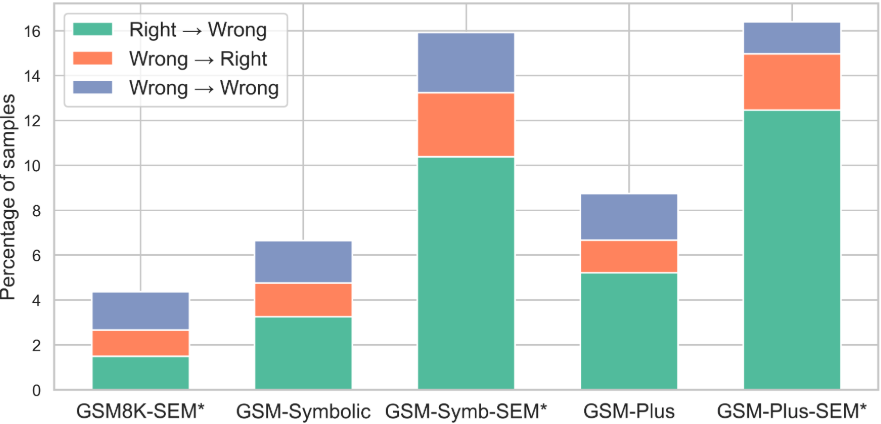}
    \caption{GSM8K vs. variant outcomes (all models pooled).}
    \label{fig:pooled}
\end{figure}

\begin{figure}[!b]
    \centering
    \includegraphics[width=0.85\linewidth]{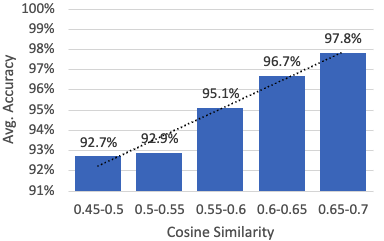}
    \caption{Average accuracy across models for GSM8K-SEM in different cosine similarity ranges w.r.t. GSM8K.}
    \label{fig:cosim}
\end{figure}

\paragraph{Comparison of Perturbation Types}Although GSM8K-SEM, GSM-Symbolic, and GSM-Plus induce accuracy drop, the underlying perturbations are fundamentally different. GSM-Symbolic primarily substitutes names or numbers, resulting in high semantic similarity to the original GSM (Figure~\ref{fig:cossimtrend}). GSM-Plus on the other hand induces perturbations through numerical and arithmetic variations, problem rephrasing, and distractor insertion, which involves injecting topic-related but irrelevant sentences that do not contribute to the solution. In contrast, GSM8K-SEM produces variants with substantial semantic divergence, despite maintaining final answer consistency, resulting in lower cosine similarity. Figure~\ref{fig:pooled} shows the difference in performance across these perturbation types in further detail. It shows across perturbations, the breakdown of samples that change from right to wrong, wrong to right, and wrong to wrong compared to GSM8K, where SEM added on top of Symbolic and Plus causes much larger differences and more failed answer determinations. Together this shows that SEM variants and other existing variants probe distinct aspects of LLM generalization.
\paragraph{Relationship to Semantic Similarity}
Figure~\ref{fig:cosim} illustrates model accuracy as a function of cosine similarity with the original GSM dataset. Accuracy generally improves with higher semantic similarity, implying that greater semantic shifts lead to larger performance drops, even when the required computations and numbers remain unchanged.

\begin{figure}[!t]
    \centering    \includegraphics[width=1\linewidth]{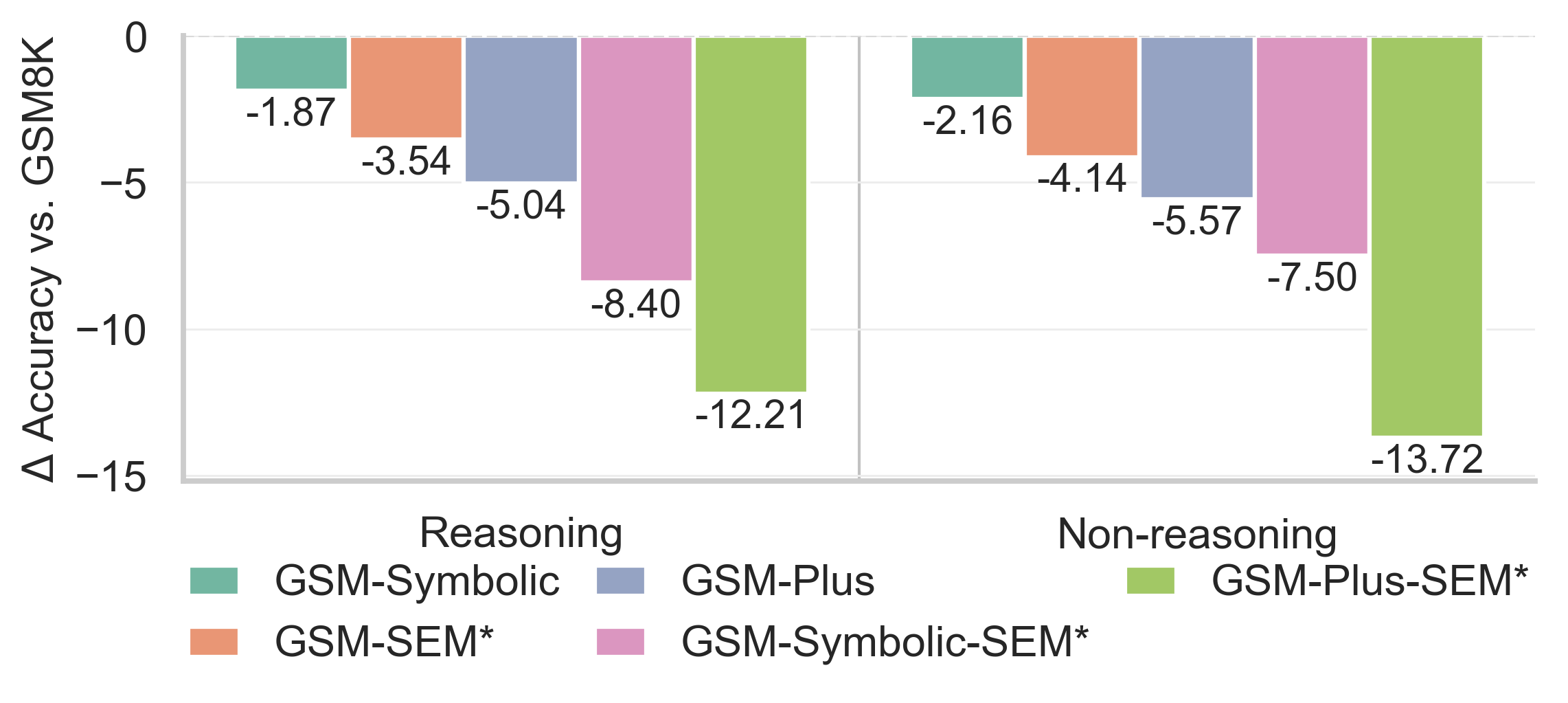}
    \caption{Delta in performance for GSM-variants compared to baseline (GSM8K) by reasoning (O3, Gemini-2.5-Pro, GPT5x) vs non-reasoning models (others).}
    \label{fig:modelfamdiff}
\end{figure}

\paragraph{Model-Specific Effects}
The perturbation challenges models, with slight variation depending on the architecture and presence of reasoning specialized ability in models. In Table~\ref{tab:gsmresdetailed}, GSM8K-SEM shows 3.61\% average degradation on GPT family of models, compared to 4.33\% on Gemini and 4.84\% on Llama family. Figure \ref{fig:modelfamdiff} shows delta in performance by reasoning vs non-reasoning models averaged by all model families tested. While reasoning-optimized models exhibit a generally smaller overall performance drop except Symbolic-SEM, the improvement over non-reasoning models is marginal. This suggests the gap is broad-based rather than confined to non-reasoning models.

\section{Analysis of samples}

\paragraph{Samples that improved under variants} Some samples that originally yielded inaccurate results showed stronger performance in the variants. These tended to be cases where original question has a strong ambiguity which the variants clarify. An example is as follows.
\begin{tcolorbox}[
  title={Example: Variants overcoming ambiguity},
  colback=gray!4, colframe=gray!55,
  colbacktitle=gray!65,   % darker title bar
  coltitle=white,         % title text color
  boxrule=0.5pt, arc=1.5mm,
  left=1.5mm,right=1.5mm,top=1mm,bottom=1mm,
  boxsep=0.6mm,
  toptitle=0.4mm,bottomtitle=0.2mm,
  fonttitle=\small\bfseries
 ]%[
%   title=Example: Variants overcoming ambiguity,
%   colback=gray!4, colframe=gray!55, boxrule=0.5pt, arc=1.5mm,
%   left=1.5mm,right=1.5mm,top=1mm,bottom=1mm
% ]
\begin{small}
\emph{Original:}
    Watson works a 10-hour shift each day, five days a week. He earns \$10 per hour and gets a \$300 bonus each week if the company performs well. How much money did Watson make in April if the company performed very well for the whole month?\\

\emph{Variant:}
    What is the total amount of money Watson earned in April, given that he makes \$10 per hour, works 10 hours a day, 5 days a week, and received a \$300 bonus for each of the \underline{4 weeks} of the month?
    
\end{small}
\end{tcolorbox}

The variant clarifies that a month should be treated as 4 weeks, whereas the original question leaves this assumption ambiguous (e.g., some may instead assume a 30-day month).

\paragraph{Negative Result Analysis.}
Across GSM-derived benchmarks, SEM variants underperform their base datasets (Table~\ref{tab:gsmresdetailed}) despite preserving the intended computation and answer. To understand why, we manually inspected items where some variants are solved reliably but other variants (from the same base problem) fail across most models, excluding uniformly always-correct/always-incorrect cases to isolate wording-driven effects.

Key failure modes: \textbf{(1) Added indirection increases cognitive load:} variants that introduce extra conversions or indirect constraints (e.g., ``\(7\%\) remaining'' vs.\ an explicit remainder) often lead models to drop or mis-bind constraints at the final step. 
\textbf{(2) Unit/currency cues trigger scale errors:} mixed ``cents''/\$/\(\textyen\) phrasing can cause implicit rescaling (e.g., treating \$1200 as \$12.00), yielding coherent-but-wrong solutions.
\textbf{(3) Reference-style rewrites destabilize state tracking:} logically equivalent relational phrasing (``twice the previous interval until the third'') is harder than direct numbers, frequently misplacing doubling/halving boundaries.
\textbf{(4) Template misretrieval under minor semantic shifts:} small cue changes can switch the latent solution template, producing structured mistakes such as, for instance, assuming available spoons for dinner table setting includes spoons already used during dinner preparation instead of counting them an unavailable due to prior usage. 
\textbf{(5) Overcomplication and ``unmotivated'' misses:} some models add unnecessary steps even for similarly simple variants, and we observe repeatable identical errors at temperature 0.0, suggesting cue-driven brittleness rather than pure decoding noise.

\begin{tcolorbox}[
  title={Example: Variants overcoming ambiguity},
  colback=gray!4, colframe=gray!55,
  colbacktitle=gray!65,   % darker title bar
  coltitle=white,         % title text color
  boxrule=0.5pt, arc=1.5mm,
  left=1.5mm,right=1.5mm,top=1mm,bottom=1mm,
  boxsep=0.6mm,
  toptitle=0.4mm,bottomtitle=0.2mm,
  fonttitle=\small\bfseries
 ]%[
%   title=Example: Direct vs.\ percent remainder,
%   colback=gray!4, colframe=gray!55, boxrule=0.5pt, arc=1.5mm,
%   left=1.5mm,right=1.5mm,top=1mm,bottom=1mm
% ]
\begin{small}
\textit{Direct remainder:}
    Arjun went to a store with \(\textyen 1100\). He bought crackers priced at \(\textyen 0.25\) each and pretzels priced at \(\textyen 1.00\) each, spending all but \(\textyen 11\) in change. If he bought 4 crackers, how many pretzels did he purchase?\\
    
\textit{Percent remainder (often harder):}
    In a market, fruit baskets are sold for 25 cents and vegetable baskets for 100 cents. Priya spent \$1100 and bought 4 fruit baskets, ending up with 1\% of her money as change. How many vegetable baskets did she purchase?

\end{small}
\end{tcolorbox}

\smallskip
\noindent\textbf{Do different models fail differently?}%for different reasons?}
We see \textit{\textbf{both shared and model-specific tendencies}}: unit/scale confusion, reference-based state drift, and target/remainder flips occur across model families, while some models like GPT5x more often overcomplicate or mis-handle narrative frames. Overall, SEM exposes a spectrum of cue-sensitive template failures, where even small semantic shifts redirect models into coherent-but-wrong reasoning paths.

\paragraph{Memorization problem.}
During quality filtering, we found items that appear ``incorrect'' to humans yet are answered correctly across models, suggesting memorization: \textbf{\textit{even when a variant changes numbers and thereby required calculations, models sometimes return a final answer that still matched the correct answer}}. (see Appendix~\ref{sec:negegmem} for example.)

Appendix~\ref{sec:negeg} contains more details and examples.

\section{Transferability to out-of-domain (non-GSM) datasets}

GSM-SEM was utilized to successfully generate semantic variants on non-GSM sets.
Our results indicate that SEM variants show degraded performance on non-mathematical datasets as well, particularly those with more complex reasoning requirements that have been released multiple years ago. In the BigBench-Hard subset (2022 release), SEM variants scored average and 5\% drops in accuracy compared to original possibly due popularity and susceptibility to memorization. 
LogicBench, released mid-2024 and less widely used than BigBench, is recent enough to have avoided widespread memorization, but not obscure enough to fully rule it out; here, performance drop was negligible. 
Finally, NLR-BIRD (very recent release from November 2025) presented minimal memorization risk, and models showed no performance drop in the SEM variants. Detailed results, prompts, and generated SEM variants examples are shared in Appendix~\ref{sec:apxood}.
\section{Conclusion}
We introduced \textbf{GSM-SEM}, a stochastic generate--validate--filter framework for constructing \emph{answer-preserving} yet semantically diverse benchmark variants. GSM-SEM enables \emph{re-runnable} evaluation, reducing reliance on static test sets that can become less diagnostic under contamination and memorization risk. %Using GSM-SEM, we release three fully human-validated datasets based on GSM8K, enabling both standalone and \emph{compositional} robustness evaluation.
Across 14 SOTA models, SEM variants consistently underperform relative to their corresponding base benchmarks, with the largest degradations observed when semantic perturbations are layered on top of Symbolic and Plus transformations. These results indicate that current models remain sensitive to semantic shifts, highlighting a gap not captured by surface-level robustness variants.
%Our results demonstrate that introducing SEM variants leads to notable performance declines across all GSM-derived datasets, with the most pronounced drops observed when semantic perturbations are combined with Symbolic or Plus variations. Revealing
%Our analysis attributes these performance drops to unexplained model confusion choosing a wrong reasoning path when semantics differ under same computations and complexity, at times over-complicating the solution, and increased cognitive load with heightened reasoning demands introduced by semantic diversity. Overall, 
%GSM-SEM offers an extensible approach for better exposing reasoning ability of language models and lowering the risks of benchmark overfitting and memorization in evaluations, offering a complementary axis of robustness beyond existing perturbations.
GSM-SEM is reusable to create diverse augmentations each time, and can be adapted to other datasets beyond GSM for more robust evaluation and variant generation.

\section*{Limitations}

GSM-SEM is most directly applicable to datasets that include a question along with a ground-truth answer and supporting reasoning information. For datasets where the reasoning path is complex or not available, GSM-SEM may only be applied partially: in one of the dual curation steps, the curation stage may be unable to generate faithful new questions from the answer alone, or the generated variants may not match the original dataset’s complexity profile if availability of reasoned solution isn’t there. As a results, such datasets may first need to be augmented with reasoning traces to make the most of GSM-SEM pipeline.
Additionally, parts of our filtering and evaluation rely on automated judging, which can introduce systematic biases despite high agreement with human labels.

In terms of potential implications and future work, while advanced models have demonstrated improved performance on general reasoning tasks, mathematics remains a particularly challenging domain. Incorporating semantically varied data during model training has the potential to promote true reasoning skills, reducing reliance on memorization. GSM-SEM may also be used to train specialized models and Reinforcement Learning systems to improve the performance of models.
Additionally, our analysis focuses on behavioral robustness and qualitative failure patterns rather than mechanistic interpretability; understanding the internal causes (activation-level) of these failures remains useful future work. 
For Extendability / modes of \emph{Strictness Filter}, the \emph{Strictness Filter} can be tightened to drop any variant that a held-out model set solves unanimously, yielding harder evaluation subset. The retained hard variants may be reused as seeds and passed through GSM-SEM to generate additional challenging variants for robustness testing. The choice of held-out models $\mathcal{M}_{\text{hold}}$ further enables targeted analyses: older model versions can surface improvements in newer models on previously inconsistent cases, while non-reasoning models can help isolate whether reasoning-oriented models benefit specifically from explicit reasoning capability on the same filtered set. 

\bibliography{custom}

\appendix
\section{Extended Related Work}
\label{sec:relatedwork}
The GSM8K dataset \citep{cobbe2021trainingverifierssolvemath} represents a foundational benchmark comprising of over 8,000 grade school mathematical world problems, partitioned into 7473 training and 1319 test samples. While these questions require only elementary arithmetic operations, recent investigations have revealed significant vulnerabilities in how LLMs approach these seemingly straightforward tasks. The widespread adoption of GSM8K has brought concerns regarding data contamination and overfitting, as evidenced by the fact that performance can fluctuate dramatically with a minor modifications \citep{mirzadehgsm}.

% These limitations have led to efforts to generate new datasets and variants. iGSM \citep{YXLA2024-gsm1} is a math dataset created through a synthetic pipeline that captures parameter dependencies in a hierarchical and graph structure. GSM-IC \cite{shi2023largelanguagemodelseasily} shows that irrelevant context can impair LLM performance, focusing on prompting techniques. Our work, however, suggests a more fundamental issue: LLMs struggle even when given multiple shots of the same question, indicating deeper challenges in problemsolving that cannot be resolved with few-shot prompting or fine-tuning on unseen distractions or variations of the same or different difficulty levels. GSM-Plus \citep{li2024gsmpluscomprehensivebenchmarkevaluating} introduces variants of GSM8K questions but lacks symbolic templates and has a fixed size and difficulty. GSM1K \citep{zhang2024carefulexaminationlargelanguage} mirrors the style and complexity of GSM8K to identify systematic overfitting in existing models, but it is not publicly available for researchers. 
\paragraph{Synthetic and Adversarial Benchmarks.}
Recent efforts have turned to synthetic data generation to produce more challenging and diverse evaluation frameworks. The iGSM \citep{ye2024physics} dataset employs a hierarchical graph-based generation process to create math word problems with controllable complexity. In iGSM, problem parameters and their dependencies are structured in graphs, enabling the construction of questions requiring up to 21 sequential operations. The framework distinguishes between direct dependencies (explicit quantitative relationships), instance dependencies (repeated structural patterns or categorical relationships), and implicit dependencies (unstated relationships requiring inference). By controlling these dependencies and the number of operations, iGSM provides fine-grained difficulty tuning and fully verifiable solutions, offering deep insights into how models handle multi-step reasoning. 

Orthogonal to synthetic complexity, other work has introduced adversarial distractions to probe LLMs’ focus. GSM-IC (Grade School Math – Irrelevant Context) by \citep{shi-freda-2023-irrelevant-context} appends semantically plausible but logically irrelevant sentences to standard GSM8K questions. This dataset reveals that even state-of-the-art LLMs can be easily distracted: model accuracy drops significantly when extraneous context is present in the problem description. For example, a simple arithmetic question padded with an irrelevant narrative or additional numbers often confuses models into erroneous reasoning. These failures occur despite the irrelevant content having no effect on the actual solution, showing that conventional few-shot prompting or fine-tuning strategies struggle to make models robust to such superficial perturbations.

"Lost in Cultural Translation" introduces variants that replace Western-centric names and objects in GSM8K with culturally diverse equivalents to reveal that LLM reasoning performance is significantly biased toward common Western lexical patterns~\citep{karim2025lostculturaltranslationllms}.

There have been several work in rephrasing data samples, on arithmetic and more diverse datasets, showing drop in scores or using rephrased data to include variation and diversity of content seen by models. \cite{maini-etal-2024-rephrasing, long2025oatrephraseoptimizationawaretrainingdata, lunardi2025robustnessreliabilitybenchmarkbasedevaluation,yang2023rethinkingbenchmarkcontaminationlanguage,safarzadeh2025evaluatingnl2sqlsql2nl,article}.

Different from dataset variations, there has been work in reasoning evaluations to dissect math reasoning gaps, such \cite{zeng2025mrgsmk} that presents evaluation benchmark for reasoning - MR-GSM8K, and ReasonAgain \cite{yu2024reasonagainusingextractablesymbolic} that uses symbolic paradigms for mathematical reasoning evaluation. \cite{yu2024metamath} presents a fine-tuned LLM to solve for mathematical reasoning. \cite{zhou2025is} proposes MathCheck, a checklist for testing task generalization and reasoning robustness.

% \paragraph{Our Contribution: GSM-SEM.}
% Our approach differs fundamentally from prior work by introducing a structured framework for generating semantic variations of the math problems. Our work presents data augmentation framework as opposed to reasoning evaluation, making it a different offering from work like MR-GSM and ReasonAgain. In contrast to GSM-Plus’s surface-level perturbations or GSM-Symbolic’s template-based entity swaps, we alter the core facts and relationships within each problem while keeping the overall answer unchanged. This means an original GSM8K question is transformed into multiple new versions that are semantically distinct, different scenarios, values, or logical relationships, yet require the same solution steps and yield the same answer. By systematically varying the underlying semantics rather than just phrasing or numeric values, we can probe whether models truly grasp the mathematical relationships or are merely overfitting to familiar storylines. There have been explorations of generation of semantic variations of data in different contexts \cite{wang-etal-2024-sss, ross-etal-2022-tailor}, our work being the first application for purposes of augmenting datasets like GSM8K for benchmarking.

%We introduce \textbf{GSM-SEM}, a novel resource comprising 7,500 base questions from GSM8K each paired with five semantically varied counterparts (for a total of 45,000 problems). Notably, across 13 state-of-the-art LLMs we evaluated, GSM-SEM induces the largest performance drop compared to all existing GSM8K-derived benchmarks.

\section{Data curation and validation set-up}
\label{sec:datasettings}

Across both \textbf{curation} steps, temperature setting of 0.5 was used. Five different LLMs were used across the curation steps, including Llama-3.3-70B-Instruct, Llama-4-Maverick, Llama-4-Scout \citep{touvron2023llamaopenefficientfoundation}, GPT-4.1, and GPT-4o \citep{gpt4o}. Also see Appendix~\ref{sec:gsm8k_sample} to see what question and answers look like in GSM8K for an example.

Full hyperparameter settings are:
\begin{itemize}
    \item temperature (temp): 0.5
    \item top\_p: 1
    \item top\_k: -1
    \item max\_tokens: 10,000
    \item frequency\_penalty: 1.0
\end{itemize}

Curation \#1 prompt is as follows:

\noindent\fbox{%
    \begin{minipage}{0.46\textwidth} % Adjust 0.48 for desired width, slightly less than 0.5 to account for frame
Given the following answer, write an appropriate question for which this answer would be correct. Make sure the question contains all specifications required to compute the answer correctly.\\
Return only the question and no additional text.\\
<special-instruction>\\

Answer: [[answer]]

Question:
    \end{minipage}%
}

<special-instruction> is  placeholder for any dataset specific instructions. For all GSM-specific variants, it is left blank.

Curation \#2 prompt is as follows:
\noindent\fbox{%
    \begin{minipage}{0.46\textwidth} % Adjust 0.48 for desired width, slightly less than 0.5 to account for frame
Given a question, return the same question changing core facts while keeping the numbers and core problem the same. \\
Return only the question and no additional text.\\
\\
Question: Susan made 100 cookies for Christmas and was going to 
equally divide them between her 6 nephews. Before Susan could package 
them, her husband snuck 4 cookies for himself. How many cookies will
each of Susan's nephews get?\\
New Question: Mala has baked 100 cupcakes for her 6 cousins to enjoy equally, but on their arrival, she finds that 4 cupcakes are spoiled. How many cupcakes will each cousin get?\\
<special-instruction>\\
\\
Question: [[question]]\\
New Question: 
    \end{minipage}%
}
\\

For this prompt as well <special-instruction> is  placeholder for any dataset specific instructions. For all GSM-specific variants, it is left blank.

For \textbf{automated quality evaluation}, we used GPT4o as LLM-judge and prompt and parameter settings are as follows.

\noindent\fbox{%
    \begin{minipage}{0.46\textwidth} % Adjust 0.48 for desired width, slightly less than 0.5 to account for frame
Given this question, does the provided answer look correct for this question? Say True if it looks correct, else say False. Don't return any extra text in your response.\\

Question: [[question]]\\
Answer: [[answer]]\\
True or False: 
    \end{minipage}%
}

Hyperparameters are set as follows.
\begin{itemize}
    \item temperature (temp): 0.01
    \item top\_p: 1
    \item top\_k: -1
    \item max\_tokens: 10,000
    \item frequency\_penalty: 1.0
\end{itemize}

LLM-as-a-Judge model for \textbf{evaluating model-produced GSM-answers} is GPT4.1 and the prompt used is as follows:

\noindent\fbox{%
    \begin{minipage}{0.46\textwidth} % Adjust 0.48 for desired width, slightly less than 0.5 to account for frame
Question: [[question]] \\
\\
The correct final answer for this question is: [[answer]] \\
\\
Does the below model generated answer also conclude the same correct final answer? \\
Return True if it contains the correct final answer, else return False. Only return True or False and no extra text. \\
\\
Model generated answer: [[generated\_answer]]] \\
\\
True or False: 
    \end{minipage}%
}
\\
\\
The parameter settings are as follows:
\begin{itemize}
    \item temperature (temp): 0.1
    \item top\_p: 1
    \item top\_k: -1
    \item max\_tokens: 1000
    \item frequency\_penalty: 0.0
\end{itemize}

Our GSM-based evaluations rely on LLM-based judging and do not include a dedicated mechanical full computation-equivalence test. For GSM-style natural-language problems, solutions can follow multiple algebraically valid reasoning paths that arrive at the same correct answer, and evaluation is therefore grounded in outcome-level and semantic consistency rather than strict structural matching. Prior work on GSM-style benchmarks has commonly used final-answer-based evaluation, with more recent approaches incorporating LLM-based assessment to better handle linguistic and reasoning variability; our approach is consistent with this line of work~\cite{yosef2026rethinkingmathreasoningevaluation, singh2026mtoscpathllmslost}.

\textbf{Generating model output}: For GSM-based samples that are passed through the 14 LLMs, we use the following inference parameters to get the model outputs:

\begin{itemize}
    \item temperature (temp): 0.01
    \item top\_p: 1
    \item top\_k: -1
    \item max\_tokens: 10000
    \item frequency\_penalty: 1.0
\end{itemize}

\subsection{Similarity threshold choice}
To ensure the integrity of our dataset and filter out redundant or highly similar entries, we employed a cosine similarity threshold of 0.85. This value is widely recognized in information retrieval and natural language processing as an optimal \textit{high-precision} cutoff for identifying near-duplicate content while minimizing semantic noise \cite{xiao2011efficient, bilenko2003adaptive}. In the context of modern vector embeddings, a threshold of 0.85 is considered a conservative benchmark that effectively isolates semantically redundant documents, ensuring that only distinct data remains for subsequent analysis \cite{gautam2024rag, rekabsaz2017exploration}. By setting the threshold at this level, we mitigate the risk of \textit{semantic drift}—where broadly related but distinct concepts are erroneously grouped together—while maintaining a robust and non-redundant representation of the baseline data, consistent with established methodologies in large-scale text filtering and similarity-based data cleaning \cite{perez2015textual, wmt2023filter}.

\subsection{Strictness filter settings}
\label{sec:sfsettings}
\begin{figure}
    \centering
    \includegraphics[width=1\linewidth]{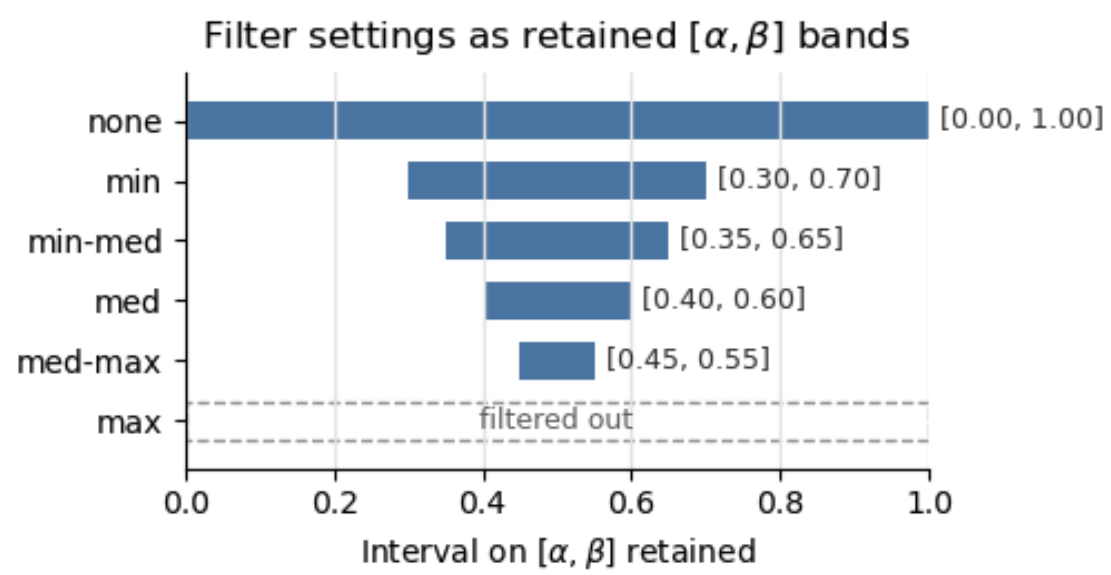}
    \caption{Strictness filter configurations. Each setting corresponds to a band on [0,1] specifying which samples (by their $[\alpha, \beta]$ values) are retained: none keeps all samples [0,1]; min–med-max retain progressively narrower central bands; max filters out all such samples.}
    \label{fig:sffiltervalues}
\end{figure}

As shown in Figure~\ref{fig:sffiltervalues}, filter configurations for strictness filter are: none (no additional filtering; all samples kept; $[\alpha, \beta]$ 0-1), min ($[\alpha, \beta]$ 0.30–0.70), min-med (0.35–0.65), med (0.40–0.60), med-max (0.45–0.55), and max (all such samples filtered out). These filters are applied to samples that yielded unanimous correctness evaluation across the held-out model set (GPT4o, GPT4o-mini, Llama-3.3-70B-Ins) in the GSM-SEM pipeline.

\section{Dataset Attributes}

\subsection{Dataset Quality Validation Methodology and Results}
\label{sec:dataqa}
To evaluate dataset quality post-collection via the SEM pipeline, we implemented a two-stage manual review process.

\begin{itemize}

\item \textbf{Validation Criteria:} Each sample was evaluated on two binary criteria:
\textbf{Quality} assessed whether the generated question correctly retained key numeric 
information from the original and preserved answer alignment. \textbf{Logical coherence} 
assessed whether the question was internally consistent and solvable (e.g., flagging 
physically implausible scenarios such as ``a shark walking on a road'').

\item \textbf{Review Process:} In \textbf{Stage 1}, 70 randomly sampled items were
evaluated by two independent professional annotators per subset, with six total reviewers
across all subsets. This stage was conducted during pipeline development to validate the
quality of the selected prompts. In \textbf{Stage 2}, every dataset sample was fully evaluated and manually reviewed to ensure a clean final dataset release. Only samples that passed human validation were included in the final dataset and used for reported results.
Each sample was independently labeled by three expert annotators, who assigned a
binary label of 1 (pass) or 0 (fail) for each criterion. Samples that did not receive unanimous
approval were additionally adjudicated through a from-scratch re-derivation of the variant.
Final inclusion labels were determined by majority vote.

\item \textbf{Metrics:} For each criterion, we report the pass rate, average pairwise
agreement, and average pairwise Cohen's $\kappa$ across the three annotator pairs.
\end{itemize}

Tables~\ref{tab:annotation_stage1} and~\ref{tab:annotation_stage2} report the results for
Stage 1 and Stage 2, respectively. The higher inter-annotator agreement in Stage 2 is
expected, as this stage was conducted after the prompt set and generation pipeline had been
refined based on Stage 1 feedback. In Stage 2, the quality criterion achieved a pass rate of
96.1\% with an average $\kappa$ of 0.725, indicating substantial agreement. Logical coherence
achieved a pass rate of 99.4\% with an average pairwise agreement of 98.1\%. The lower
$\kappa$ of 0.279 for logical coherence is attributable to the well-documented prevalence
effect: when labels are heavily skewed toward one class, Cohen's $\kappa$ can underestimate
true agreement even when raw agreement is high. We therefore report raw agreement
alongside $\kappa$ to provide a more complete picture. For the `min' strictness filter setting,
for example, 20 of the initial 518 GSM8K-SEM samples (3.9\%) were rejected: 17 failed only
the quality criterion and 3 failed both criteria, leaving 498 samples in the final set.

\begin{table}[h]
\centering
\begin{tabular}{lcc}
\hline
Metric & Quality & Logical Coherence \\
\hline
Pass Rate       & 96.4\%  & 97.9\% \\
Agreement       & 95.7\%  & 95.7\% \\
Cohen's $\kappa$ & 0.379  & 0.000$^\dagger$ \\
\hline
\end{tabular}
\caption{Stage 1 annotation results on 70 randomly sampled items.
$^\dagger$\,Cohen's $\kappa$ of 0.000 reflects the prevalence effect: 
with very few negative cases, $\kappa$ is artificially deflated despite 
high raw agreement.}
\label{tab:annotation_stage1}
\end{table}

\begin{table}[h]
\centering
\begin{tabular}{lp{1.5cm}p{2cm}}
\hline
Metric         & Quality & Logical Coherence \\
\hline
Pass Rate                & 96.1\%  & 99.4\% \\
Avg Agreement & 98.3\%  & 98.1\% \\
Avg Cohen's $\kappa$   & 0.725   & 0.279$^\dagger$ \\
\hline
\end{tabular}
\caption{Stage 2 annotation results on all GSM8K-SEM samples.
$^\dagger$\,See prevalence effect note in Table~\ref{tab:annotation_stage1}.}
\label{tab:annotation_stage2}
\end{table}

\subsection{Dataset size attributes}
\label{sec:datasetsize}
In this work, we present our results alongside those obtained using the GSM-Symbolic and GSM-Plus. To ensure a fair comparison, we utilized the same 100 GSM8K samples for which the GSM-Symbolic dataset was initially released. For each of these 100 samples, we generated multiple augmentations to produce the GSM8K-SEM dataset, resulting in a total of 685 samples. The \emph{Semantic Variant Generation} was prompted 10 times to produce 10 variants per sample. This yielded 1000 samples. The Validation Stage removed ~20\% of the samples, and Redundancy Pruning stage removed ~10\% of the samples left, and an additional ~3.9\% samples were removed after 100\% human assessment, yielding 685 samples for `none' \emph{Strictness Filter} setting. See Table~\ref{tab:variant-strictness-sizes} for numbers on other settings.

We further applied the SEM augmentation process to GSM-Symbolic variants. The GSM-Symbolic dataset provides 50 variations for each of the 100 original GSM8K samples. From these, we randomly selected 5 GSM-Symbolic variants per sample, giving us 100×5=500 GSM-Symbolic base variants. For each of these 500 variants, we then created multiple GSM-SEM-augmented versions, leading to several augmented samples per variant (across the datasets, the \emph{Semantic Variant Generation} was prompted 10 times to produce 10 variants per input sample, here 500 x 10 = 5000 samples, then filtered down by various pipeline steps down to numbers shared in Table~\ref{tab:variant-strictness-sizes}). This approach produced multiple variations for each of the 100 original GSM8K samples, in line with the GSM-Symbolic methodology.
In total, our GSM-Symbolic-SEM dataset contains 2525 samples, derived from the SEM augmentation of selected GSM-Symbolic variants based on the original GSM8K test data. For GSM-Plus, we used the same base 100 original samples to get all variants across their 8 variant types. We exclude the critical thinking variant type because it requires evaluating whether a model asks follow-up questions when the prompt is underspecified. We omit it to remain consistent with the other benchmarks we use, which are single-turn tasks evaluated by final answer / calculation correctness, rather than by interactive follow-up behavior.

Using the same 100 base GSM8K samples, we generated 1436 samples from GSM-Plus.

For GSM-paraphrased, we paraphrased each of the 100 base GSM8K samples multiple times, leading to a total of ~500 samples of which manual inspection deemed 14 samples as low quality (changed problem/calculation needed after paraphrasing), so we excluded those and our final dataset size was 685.

All results are first averaged across the variants for each of the 100 original GSM8K samples, yielding a single accuracy score per sample. These 100 per-sample accuracy scores are then averaged to produce the final, overall accuracy for the dataset. In this way, each original sample contributes equally to the final metric, regardless of the number of its variants.

In terms of dataset size based on the different \emph{Strictness filter}, Table~\ref{tab:variant-strictness-sizes} shows the breakdown of the size.

\begin{table}[!t]
\centering
\begin{tabular}{p{1.49cm}p{1.3cm}p{1.5cm}p{1cm}}
\hline
Strictness Filter & \multicolumn{3}{c}{Data Size} \\
\hline
& GSM8K-SEM & GSM-Symbolic-SEM & GSM-Plus-SEM\\
\hline
none   & 685 & 2525 & 1436 \\
min    & 498 & 1866 & 981 \\
min-med & 386 & 1513 & 802 \\
med    & 300 & 1150 & 637 \\
med-max & 199 & 791 & 456 \\
max    & 111 & 391 & 279 \\
\hline
\end{tabular}
\caption{Dataset sizes by variant and strictness filter.}
\label{tab:variant-strictness-sizes}
\end{table}

\subsection{Dataset analysis - cosine similarity}
\label{sec:cosine}

\begin{figure}
    \centering
    \includegraphics[width=1\linewidth]{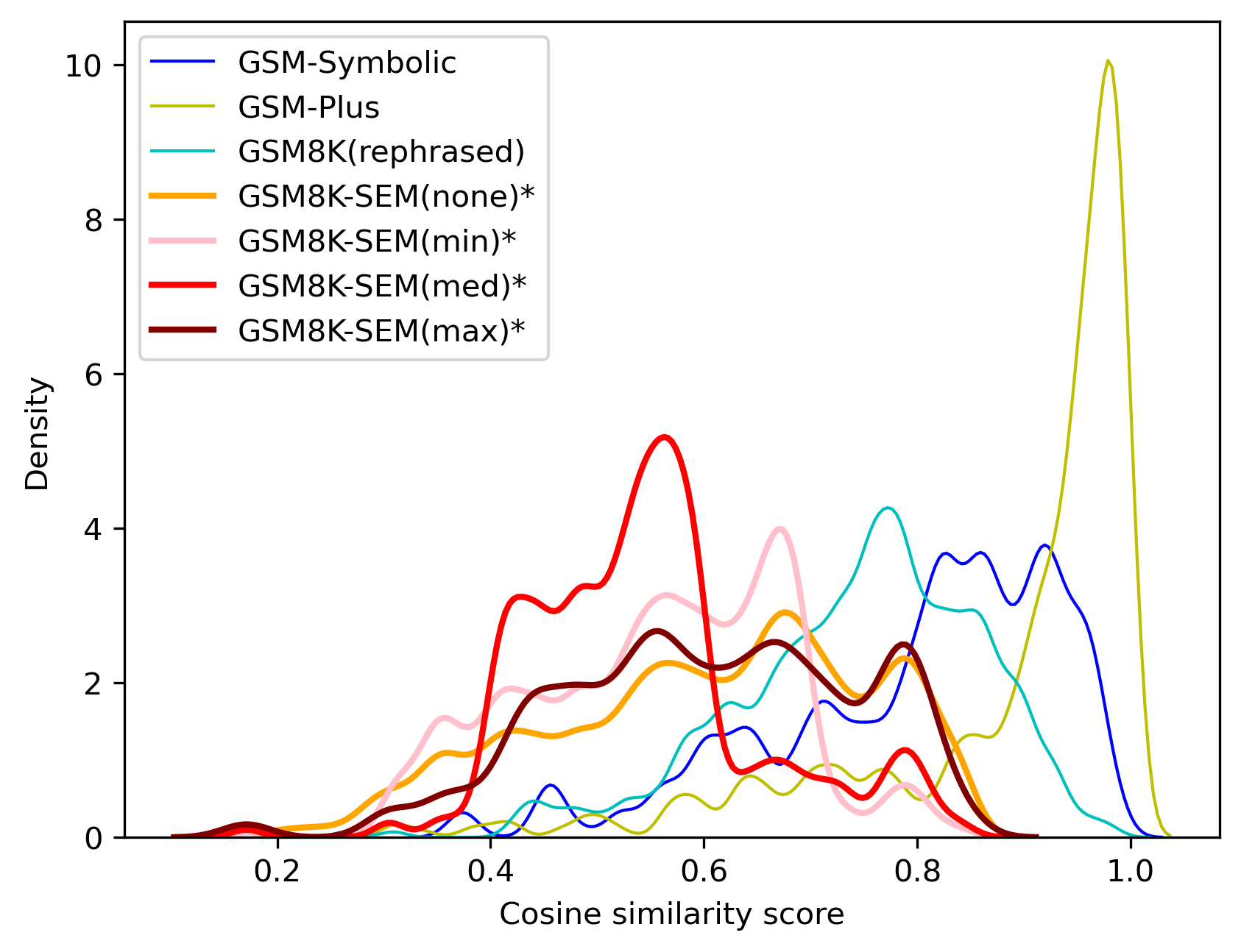}
    \caption{Cosine similarity distribution of GSM8K variants with respect to GSM8K for different \emph{Strictness Filter} configurations. Filter settings: none (no additional filtering; all samples kept), min ($[\alpha,\beta]$ 0.30–0.70), med (0.40–0.60), and max (all such samples filtered out). The med set, for instance, shows a concentration around mid-range cosine values as the filtering removes only certain extremes (rather than uniformly pruning across cosine similarity ranges). By contrast, the max set is filtered primarily by held-out model unanimity across a wider range of cosine similarities, which yields a more uniform cosine distribution.}
    \label{fig:cossimSFall}
\end{figure}

\begin{figure}
    \centering
    \includegraphics[width=1\linewidth]{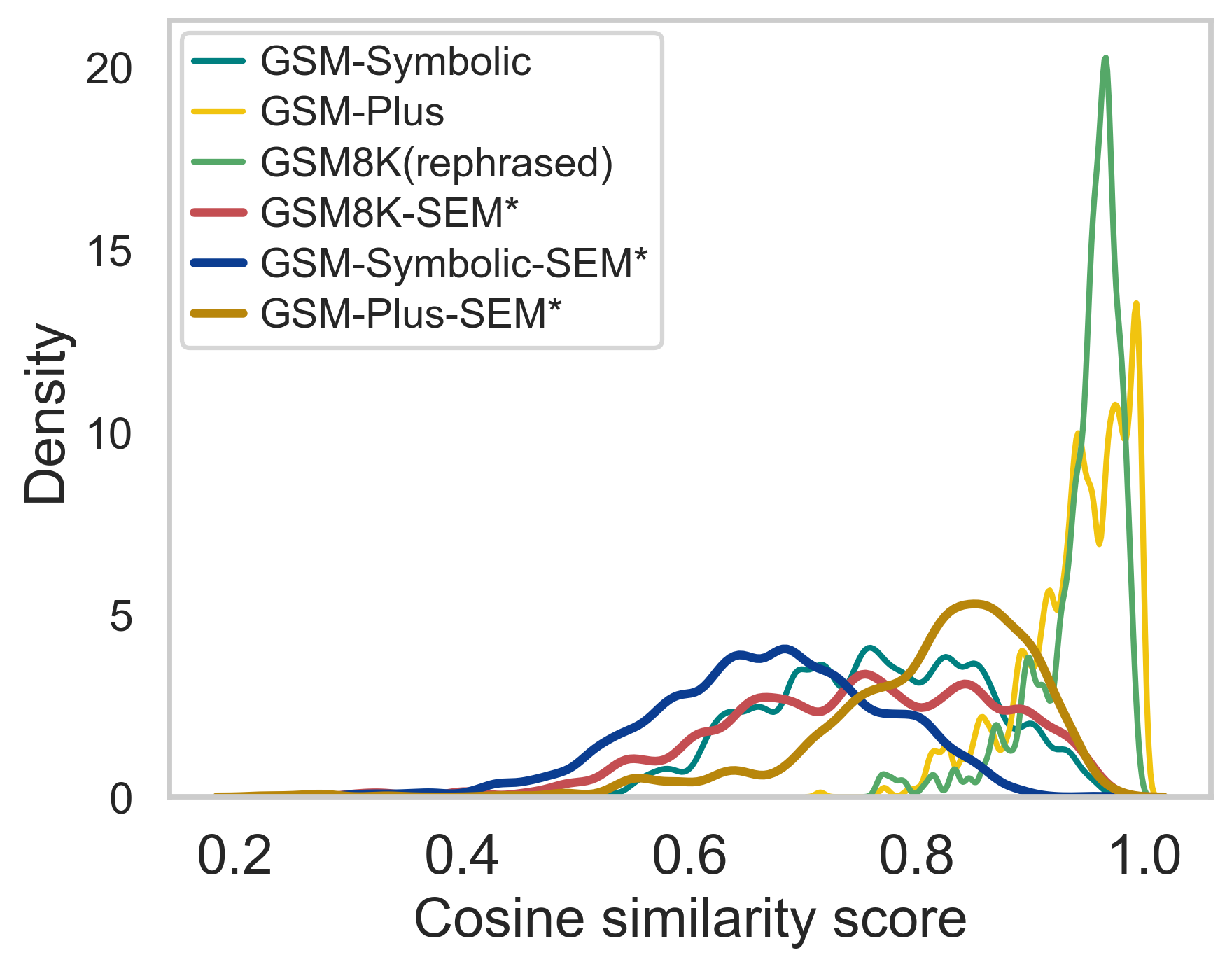}
    \caption{Cosine similarity distribution using all-MiniLM-L6-v2 embeddings.}
    \label{fig:cossimbert}
\end{figure}

We also compare the cosine similarity distribution using a sentence transformer model (all-MiniLM-L6-v2) generated embeddings over the same sets to  measure their semantic closeness to the original questions, as shown in Figure~\ref{fig:cossimbert}. From a semantic perspective, paraphrased questions denote the same set of propositions as the original ones, hence the highest similarities among the four sets. The symbolic semantic set, generated via two steps of transformations: entity swap followed by semantic variations, understandably, has the lower similarities with the original. GSM-Symbolic largely overlaps with GSM8K-SEM, but has a wider spread. As the number of entities in each query varies, it adds to the similarity variations among the GSM-Symbolic set.

\section{Results - Chain of Thought}
\label{sec:cot}
Some existing literature~\cite{mirzadehgsm} presents results on GSM8K and its variants using 8-cot (chain of thought) prompts. We conducted experiments using 8-cot prompts as well and noticed high alignment with results using zero-shot prompts. Table~\ref{tab:cot} shows 8-cot prompt results across GSM and some of its variants in comparison to zero-shot prompts. The difference in results was marginal, well under than 1\%.

\begin{table}[h]
\centering
\begin{tabular}{lccc}
      & 8-cot & zero-shot & |diff| \\
      \hline
GSM8K & 89.57\%      &               89.86\%  &               0.29\%\\
Paraphrased   & 87.06\%      & 87.57\%         & 0.51\%\\
GSM-Symbolic     & 87.09\%      & 86.23\%  & 0.86\%       \\
\hline
\end{tabular}
%}
\caption{Accuracy across 9 LLMs for 8-cot prompt and zero-shot prompt against GSM and its variants.}
\label{tab:cot}
\end{table}

\section{Results by Model}
\label{sec:results}

% \begin{figure*}
%     % \centering
%     \includegraphics[width=1\linewidth]{eval-models-datasets.png}
%     \caption{Model accuracy across GSM variants and competitive models.}
%     \label{fig:eval}
% \end{figure*}

%Figure~\ref{fig:evaldrops} shows the accuracy drop by model across GSM-SEM variants.% and Figure~\ref{fig:eval} shows results for GSM, GSM-paraphrased, GSM-Symbolic, GSM-SEM, and GSM-Symbolic-SEM broken down by LLM.

Results in this paper are shown using (1) the original GSM benchmark using the 100 samples for which GSM-Symbolic released their augmented dataset, compared against (2) paraphrased versions of the original GSM questions, (3) the GSM-Symbolic benchmark, (4) our new GSM8K-SEM benchmark applied to the original GSM 100 samples, (5) GSM-Symbolic-SEM, where our SEM augmentation pipeline is applied on top of the GSM-Symbolic dataset, and (6) GSM-Plus-SEM, where our SEM augmentation pipeline is applied on top of the GSM-Plus dataset for the same baseline 100 GSM8K samples. The results demonstrate that accuracy declines as semantic variation increases, with the largest drop observed on the GSM-SEM-augmented benchmarks.% * indicates results generated by our approach. 

Table~\ref{tab:gsmresdetailed} and Figure~\ref{fig:respipelinethrs} in main paper contains results and expanded results are shared in Table~\ref{tab:strictness_drops_sem},~\ref{tab:symbolic_sem_strictness_drops},~\ref{tab:plus_sem_drops}, and ~\ref{tab:pvalues-by-model}.

\subsection{Results by Model across Strictness Filter configurations}

Drop in accuracy from GSM8K-SEM is shared in Table~\ref{tab:strictness_drops_sem}, GSM-Symbolic-SEM in Table~\ref{tab:symbolic_sem_strictness_drops}, and GSM-Plus-SEM in Table~\ref{tab:plus_sem_drops}.

\begin{table*}[t]
\centering
\small
\begin{tabular}{p{2cm}p{1.5cm}p{1.5cm}p{1.5cm}p{1.5cm}p{1.5cm}p{1.5cm}}
\hline
 & \multicolumn{6}{c}{GSM8K-SEM} \\
\cline{2-7}
 & none & min & min-med & med & med-max & max \\
\hline
Grok3           & -1.09\% (0.1652) & -1.78\% (0.0752) & -2.59\% (0.0464)& -3.70\% (0.0398)& -6.36\% (0.0108)& -13.96\% (0.0035)\\
Llama3.1-405    & -4.75\% (<1E-4) & -5.20\% (0.0001) & -5.93\% (<1E-4)& -6.84\% (<1E-4)& -8.89\% (<1E-4)& -13.49\% (0.0001)\\
Llama4-Mav      & -1.35\% (0.0139) & -1.89\% (0.0197) & -2.79\% (0.0176)& -3.79\% (0.0138)& -5.52\% (0.0108)& -12.40\% (0.0065)\\
Llama4-Scout    & -1.40\% (0.0346) & -2.52\% (0.0088)& -3.26\% (0.0081)& -3.88\% (0.0124)& -5.35\% (0.0114)& -12.72\% (0.0032)\\
GPT4.1          & -2.81\% (0.0353) & -3.20\% (0.0321)& -3.25\% (0.0353)& -3.85\% (0.0247)& -5.47\% (0.0267)& -12.19\% (0.0115) \\
GPT4.1-mini     & -1.70\% (0.0098) & -1.76\% (0.0197) & -2.11\% (0.0197) & -2.68\% (0.0156)& -3.45\% (0.0156)& -6.43\% (0.0192)\\
Gemini2.5-f     & -3.57\% (0.0019) & -3.94\% (0.0039) & -4.52\% (0.0055) & -5.07\% (0.0055)& -8.02\% (0.0039)& -14.41\% (0.0064)\\
Gemini2.5-f-l   & -1.77\% (0.0517) & -2.82\% (0.0240) & -3.50\% (0.0357)& -4.44\% (0.0246) & -6.38\% (0.0100)& -17.32\% (0.0013)\\
Gemini2.5-pro   & -2.23\% (0.0070) & -2.32\% (0.0095) & -2.88\% (0.0108)& -3.49\% (0.0064) & -6.22\% (0.0009)& -12.70\% (0.0021)\\
O3              & -1.97\% (0.0070) & -2.74\% (0.0012)& -3.01\% (0.0009)& -3.36\% (0.0011)& -4.64\% (0.0009)& -8.82\% (0.0029)\\
GPT5(mnml)      & -2.21\% (0.0183) & -2.77\% (0.0025) & -3.00\% (0.0025)& -3.37\% (0.0025)& -4.81\% (0.0025)& -8.00\% (0.0090)\\
GPT5.1(mnml)    & -2.93\% (0.0009)& -3.40\% (0.0011) & -3.70\% (0.0024)& -4.70\% (0.0018)& -6.77\% (0.0015)& -15.36\% (0.0002)\\
GPT5(med)       & -1.42\% (0.2382) & -1.94\% (0.1010) & -2.21\% (0.0566) & -2.59\% (0.0458)& -3.85\% (0.0293)& -6.65\% (0.0176)\\
GPT5.1(med)     & -2.82\% (0.0069) & -3.22\% (0.0079) & -3.62\% (0.0078)& -4.47\% (0.0024)& -6.34\% (0.0024)& -13.54\% (0.0006) \\
\hline
Avg             & -2.29\% & -2.82\% & -3.31\% & -4.02\% & -5.86\% & -12.00\% \\
Stat sig (models w/ p<0.05) & 11/14 & 12/14 & 13/14& 14/14& 14/14&14/14\\
\hline
\end{tabular}
\caption{Accuracy $\Delta$ (p-value from Wilcoxon test) across strictness filter settings in GSM8K-SEM.}
\label{tab:strictness_drops_sem}
\end{table*}

\begin{table*}[t]
\centering
\small
\begin{tabular}{p{2cm}p{1.5cm}p{1.5cm}p{1.5cm}p{1.5cm}p{1.5cm}p{1.5cm}}
\hline
 & \multicolumn{6}{c}{GSM-Symbolic-SEM} \\
\cline{2-7}
 & none & min & min-med & med & med-max & max \\
\hline
Grok3           & -2.05\% (0.0045) & -3.74\% (0.0042) & -4.49\% (0.0046)& -7.11\% (0.0019)& -10.81\% (0.0007)& -26.09\% (<1E-4)\\
Llama3.1-405    & -6.19\% (<1E-4)& -7.28\% (<1E-4)& -8.20\% (<1E-4)& -9.73\% (<1E-4) & -11.39\% (<1E-4)& -25.05\% (<1E-4)\\
Llama4-Mav      & -4.27\% (<1E-4)& -5.15\% (<1E-4)& -5.90\% (<1E-4)& -7.34\% (<1E-4)& -10.46\% (<1E-4)& -23.48\% (<1E-4)\\
Llama4-Scout    & -4.32\% (<1E-4)& -5.50\% (<1E-4)& -6.20\% (<1E-4)& -7.97\% (<1E-4)& -11.51\% (<1E-4)& -28.68\% (<1E-4)\\
GPT4.1          & -3.29\% (0.0003)& -4.48\% (0.0003)& -5.50\% (0.0002) & -7.98\% (0.0001)& -10.89\% (<1E-4)& -27.67\% (<1E-4)\\
GPT4.1-mini     & -3.61\% (<1E-4) & -4.63\% (<1E-4)& -5.21\% (<1E-4)& -6.83\% (<1E-4)& -9.28\%  (<1E-4)& -23.46\% (<1E-4)\\
Gemini2.5-f     & -2.64\% (0.0002) & -3.73\% (0.0003) & -4.61\% (0.0002) & -6.97\% (0.0002)& -11.09\% (<1E-4)& -25.77\% (<1E-4)\\
Gemini2.5-f-l   & -2.67\% (0.0001) & -4.21\% (0.0001)& -5.20\% (0.0001) & -7.27\% (0.0001) & -10.07\% (0.0001)& -23.66\% (<1E-4)\\
Gemini2.5-pro   & -3.31\% (0.0001) & -4.14\% (0.0001) & -4.74\% (0.0001) & -6.66\% (0.0001)& -9.42\% (0.0001) & -25.91\% (<1E-4)\\
O3              & -4.34\% (<1E-4) & -5.52\% (<1E-4) & -6.08\% (<1E-4)& -8.71\% (<1E-4)& -11.67\% (<1E-4)& -24.26\% (<1E-4)\\
GPT5(mnml)      & -4.40\% (<1E-4)& -5.49\% (<1E-4)& -6.18\% (0.0001) & -8.68\% (<1E-4)& -11.63\% (<1E-4)& -22.36\% (<1E-4)\\
GPT5.1(mnml)    & -5.73\% (<1E-4)& -7.62\% (<1E-4)& -8.32\% (<1E-4)& -11.26\% (<1E-4)& -15.64\% (<1E-4)& -34.45\% (<1E-4)\\
GPT5(med)       & -3.01\% (0.0002) & -4.37\% (0.0002) & -4.94\% (0.0002) & -7.60\% (0.0001)& -11.09\% (<1E-4)& -24.65\% (<1E-4)\\
GPT5.1(med)     & -6.39\% (<1E-4)& -8.27\% (<1E-4)& -8.87\% (<1E-4)& -11.80\% (<1E-4)& -14.79\% (<1E-4)& -31.92\% (<1E-4)\\
\hline
Avg.             & -4.02\% & -5.30\% & -6.03\% & -8.28\% & -11.41\% & -26.24\% \\
Stat sig (models w/ p<0.05) & 14/14 & 14/14 & 14/14 & 14/14 & 14/14 & 14/14 \\
\hline
\end{tabular}
\caption{Accuracy $\Delta$ (p-value from Wilcoxon test) across strictness filter settings in GSM-Symbolic-SEM.}
\label{tab:symbolic_sem_strictness_drops}
\end{table*}

\begin{table*}[t]
\centering
\small
\begin{tabular}{p{2cm}p{1.5cm}p{1.5cm}p{1.5cm}p{1.5cm}p{1.5cm}p{1.5cm}}
\hline
 & \multicolumn{6}{c}{GSM-Plus-SEM} \\
\cline{2-7}
 & none & min & min-med & med & med-max & max \\
\hline
Grok3           & -3.72\% (<1E-4) & -7.34\% (<1E-4)  & -9.83\% (<1E-4)& -13.00\% (<1E-4)& -17.49\% (<1E-4)& -29.19\% (<1E-4)\\
Llama3.1-405    & -6.66\% (<1E-4) & -9.75\% (<1E-4) & -11.43\% (<1E-4)& -13.06\% (<1E-4)& -17.10\% (<1E-4)& -27.09\% (<1E-4)\\
Llama4-Mav      & -5.74\% (<1E-4) & -9.16\% (<1E-4) & -11.98\% (<1E-4)& -16.34\% (<1E-4)& -19.97\% (<1E-4)& -32.21\% (<1E-4)\\
Llama4-Scout    & -4.96\% (<1E-4) & -8.32\% (<1E-4)  & -11.36\% (<1E-4)& -15.33\% (<1E-4) & -19.17\% (<1E-4)& -30.90\% (<1E-4)\\
GPT4.1          & -4.97\% (<1E-4) & -8.38\% (<1E-4)  & -10.35\% (<1E-4)& -11.86\% (<1E-4)& -16.10\% (<1E-4) & -27.13\% (<1E-4) \\
GPT4.1-mini     & -3.42\% (<1E-4) & -6.05\%  (<1E-4) & -8.58\% (<1E-4) & -12.19\% (<1E-4)& -14.69\% (<1E-4)& -24.87\% (<1E-4)\\
Gemini2.5-f     & -5.91\% (<1E-4) & -9.32\% (<1E-4)  & -11.50\% (<1E-4)& -13.38\% (<1E-4)& -16.91\% (<1E-4)& -26.13\% (<1E-4)\\
Gemini2.5-f-l   & -4.70\% (<1E-4) & -8.71\% (<1E-4)  & -11.66\% (<1E-4)& -14.05\% (<1E-4)& -17.09\% (<1E-4)& -27.45\% (<1E-4)\\
Gemini2.5-pro   & -5.12\% (<1E-4)& -8.14\% (<1E-4) & -9.86\% (<1E-4) & -11.53\% (<1E-4)& -15.19\% (<1E-4)& -25.47\% (<1E-4)\\
O3              & -3.41\% (0.0001) & -6.00\% (<1E-4)  & -8.05\%  (<1E-4)& -9.77\% (<1E-4) & -12.76\% (<1E-4)& -21.47\% (<1E-4)\\
GPT5(mnml)      & -5.70\% (<0.0001) & -8.27\%  (<1E-4)& -10.14\% (<1E-4)& -12.86\% (<1E-4) & -15.43\% (<1E-4)& -25.64\% (<1E-4)\\
GPT5.1(mnml)    & -6.09\% (<1E-4) & -10.27\% (<1E-4)& -13.73\% (<1E-4)& -18.52\% (<1E-4)& -21.82\% (<1E-4)& -39.04\% (<1E-4)\\
GPT5(med)       & -3.45\% (0.0004) & -5.85\% (0.0005) & -7.24\% (0.0004) & -9.09\% (0.0002)& -11.52\% (0.0001)& -19.35\% (<1E-4)\\
GPT5.1(med)     & -6.79\% (<1E-4) & -11.11\% (<1E-4) & -14.03\% (<1E-4)& -18.40\% (<1E-4)& -20.33\% (<1E-4)& -31.56\% (<1E-4) \\
\hline
Avg.           & -5.04\% & -8.33\%  & -10.70\% & -13.53\% & -16.83\% & -27.68\% \\
Stat sig (models w/ p<0.05) & 14/14 & 14/14& 14/14& 14/14& 14/14&14/14\\
\hline
\end{tabular}
\caption{Accuracy $\Delta$ (p-value from Wilcoxon test) across strictness filter settings in GSM-Plus-SEM.}
\label{tab:plus_sem_drops}
\end{table*}

\subsection{Statistical Analysis - detailed}
\label{sec:statdetailed}

Table~\ref{tab:pvalues-by-model} contains p-values from a Wilcoxon statistical test per model and per variant set for paraphrased version of GSM8K, GSM-Symbolic, and GSM-Plus. The same statistic across strictness filter configurations in SEM variants for GSM8K-SEM is shared in Table~\ref{tab:strictness_drops_sem}, GSM-Symbolic-SEM in Table~\ref{tab:symbolic_sem_strictness_drops}, and GSM-Plus-SEM in Table~\ref{tab:plus_sem_drops}. The test measures if drop in performance compared to GSM8K is statistically significant.

\begin{table}[htbp]
\centering
\begin{tabular}{p{2.3cm}p{1.2cm}p{1.2cm}p{1.2cm}}
\hline
 & Para-phrased & GSM-Symbolic & GSM-Plus  \\
\hline
Grok3 & 0.954  & 0.429 & 0.001  \\
Llama3.1-405 & 0.587  & 0.005  & 0.000  \\
Llama4-Mav & 0.063  & 0.002  & 0.000  \\
Llama4-Scout & 0.153  & 0.042  & 0.000  \\
GPT4.1 & 0.672  & 0.035 & 0.000 \\
GPT4.1-mini & 0.672  & 0.035  & 0.000  \\
Gemini2.5-f & 0.142  & 0.005  & 0.000  \\
Gemini2.5-f-l & 0.225 & 0.288  & 0.000  \\
Gemini2.5-pro & 0.664  & 0.300  & 0.000  \\
O3 & 0.644  & 0.050  & 0.000  \\
GPT5(mnml) & 0.417 & 0.166  & 0.000  \\
GPT5.1(mnml) & 0.142  & 0.098  & 0.000  \\
GPT5(med) & 0.854 & 0.264  & 0.009  \\
GPT5.1(med) & 0.077 & 0.005  & 0.000  \\
\hline
\#models with p-value$<0.05$ (out of 14)
& 0 & 7 & 14 \\
\hline
\end{tabular}
\caption{P-values by model and variant set (Paraphrased, GSM-Symbolic, GSM-Plus) from Wilxocon test comparing drop in variant performance compared to GSM8K.}
\label{tab:pvalues-by-model}
\end{table}

% \section{Dataset Analysis}
% \label{sec:dataanalysis}
% For samples that consistently perform 100\% accurately on original GSM  

% - 36 out of 100 samples achieved 100\% accuracy across 13 state-of-the-art models across several runs against original GSM samples. 

% - Rephrased variants: 21 out of 36 had below 100\% accuracy 

% - Symbolic (prior art): 30 out of 36 had below 100\% accuracy 

% - (ours) 
%     - SEM: 31 samples had below 100\% accuracy, with 27 overlapping between Symbolic and SEM, where both had <100\% accuracy. 
    
%     - Symbolic-SEM: 35 out of 36 samples had below 100\% accuracy. 
    
% - Avg accuracy: \\
%     - Rephrased (prior art) variants: 97.1\% \\
%     - Symbolic (prior art) variants: 94.1\% \\
%     - SEM variants (ours): 94.8\%. \\
%     - Symbolic-SEM variants (ours): 91.7\% \\

% For samples that consistently perform 100\% accurately on original GSM and its rephrased variants across 13 SOTA models \\
% - 15 out of 100 samples achieved 100\% accuracy across 13 SOTA models in both original and rephrased variants of GSM. \\
% - Symbolic (prior art): 10 out of 15 had below 100\% accuracy for Symbolic \\
% - (ours) \\
%     - SEM: 10 had below 100\% accuracy for SEM, with the overlap of 7 samples between the SEM and Symbolic. \\
%     - Symbolic-SEM: 14 out of 15 samples had below 100\% accuracy \\
% - Avg accuracy: \\
%     - Symbolic (prior art) variants: 97.7\% \\
%     - SEM variants (ours): 97.5\% \\
%     - Symbolic-SEM variants (ours): 94.5\% \\

\section{Data analysis examples}
\label{sec:negeg}

\subsection{Problems with existing baselines}

We found many baseline questions were not logically coherent or complete in information.
For example, in GSM-Symbolic,

'Andre caught 6 cods last Sunday, the first cod he caught weighs 54 kgs, the second cod he caught weighs 46 kgs, and the last cod he caught weighs 23 kgs. If a kg of cod costs €2.0, how much will he earn after selling all the cods to the market?'

There is no information about the 3rd,4th or 5th cod. 

'Faisal caught 4 cods last Friday, the first cod he caught weighs 47 kgs, the second cod he caught weighs 55 kgs, and the last cod he caught weighs 24 kgs. If a kg of cod costs ¥1.0, how much will he earn after selling all the cods to the market?'

These questions lacked completeness.

\subsection{Unexplained answer arrival}
\label{sec:negegmem}
For the sample baseline Q and SEM variant (an incorrect variant): 

\begin{quote}Baseline: \textit{"I have 15 liters of peach drink that are \underline{four-sixths water} and I wish to add it to 20 liters of grape drink that is two-fifths water. But as I pour it, I spill 6 liters of the peach drink. How much water is in the remaining 29 liters?"}\end{quote}

\begin{quote} Variant: \textit{"A grape drink contains 40\% water, and a peach drink contains \underline{60\% water}. If you mix 20 liters of the grape drink with 15 liters of the peach drink and then accidentally spill 6 liters of the peach drink, how many liters of water are there in total in the remaining mixture?"}\end{quote}

Llama-4-maverick response is as follows.

\begin{small}
A grape drink contains 40\% water, and a peach drink contains 60\% water. If you mix 20 liters of the grape drink with 15 liters of the peach drink and then accidentally spill 6 liters of the peach drink, how many liters of water are there in total in the remaining mixture?
There are 20 x 2/5 = 8 liters of water from the 20 liters grape drink.
After 6 liter of peach drink was spilled, there were 15 - 6 = 9 liters of peach drink left.
Out of the 9 liters, 9 x 2/3 = 6 liters are water.
Thus, there are a total of 8 + 6 = 14 liters of water out of the 29 liters.
\#\#\#\# 14
\#\# Step 1: Calculate the amount of water in the grape drink.
First, we need to find out how much water is in the 20 liters of grape drink. Since it contains 40\% water, we can calculate this by multiplying 20 liters by 40\% (or 0.4).

\#\# Step 2: Perform the calculation for the amount of water in the grape drink.
Water in grape drink = 20 * 0.4 = 8 liters.

\#\# Step 3: Calculate the initial amount of water in the peach drink before spilling.
The peach drink contains 60\% water. We have 15 liters of it initially, so we need to find out how much water is in these 15 liters.

\#\# Step 4: Perform the calculation for the initial amount of water in the peach drink.
Water in peach drink = 15 * 0.6 = 9 liters.

\#\# Step \#\#5: Determine how much peach drink is left after spilling some.
6 liters of the peach drink are spilled, so we subtract this from the initial amount to find out how much is left.

\#\# Step \#\#6: Perform the calculation for remaining peach drink after spilling.
Remaining peach drink = Initial peach drink - Spilled peach drink = (15 - 6) liters = (15 - (15*6/15)) is not needed as we directly know that we spill a portion directly and not a percentage, so it's simply subtracting from total volume and then finding its percentage contribution later on.

\#\# Step \#\#7: Correctly calculate remaining volume and composition after spill.
After mixing and before spilling, total mixture volume is (20 + 15) = 35 liters. After spilling, it becomes (35 - part that was spilled). The part that was spilled was only from peach which means we're removing some portion directly from total mixture but specifically it's just removing part of one component first then considering its effect on whole mixture indirectly.

\#\# Step \#\#8: Simplify step \#\#7 by directly calculating remaining volumes and their compositions after spill correctly.
The actual step should be calculating how much liquid remains after mixing and then after spilling some. So first mix then spill:
- Total mixed before spill = Grape + Peach = (20 +15) litres
- Water content before spill:
- From Grape Drink
- From Peach Drink

Then adjust according to what was actually asked which involves understanding what happens when you "spill" some liquid.

\#\# Step \#\#9: Recalculate with correct understanding:
First mix both drinks together:
Total Volume Before Spill = Grape Drink + Peach Drink
= (20 +15)
=35 litres

Then consider what happens when you "spill" some liquid:
Spill occurs only from "peach" but since they are mixed we'll consider overall effect on final concentration or quantity directly rather than individual components separately at this stage because our main task revolves around final quantity post-spill.

\#\# Step \#\#10: Simplify steps further into direct calculations needed for final answer:
1. Water content from Grape Drink
2. Water content from Peach Drink initially
3. Adjusting quantities post-spill

Let's simplify into actual required steps now.

The best answer is calculated as follows:

1. **Water** **in** **Grape** **Drink**: $8$ litres ($40\%$ * $20$)
2. **Water** **in** **Peach** **Drink**: $9$ litres ($60\%$ * $15$)

Total Water Before Spill $=8+9=17$

Total Mixture Before Spill $=35$

After Spilling $6L$, Total Mixture Becomes $=29L$

Since we're removing a portion that's part-peach \& part-grape due to mixing; let's simplify our thought process focusing on what's being asked i.e., 'total' 'water' 'content'.

To do that accurately without overcomplicating things:

1st Find out proportion or quantity being removed \& apply accordingly.

The final answer is $\boxed{14}$.
\end{small}

\section{Out-of-domain experiment details and results}
\label{sec:apxood}

Table~\ref{tab:dataset_variant_accuracy} shows results for the out-of-domain non-GSM datasets across Llama-4-Scout and Gemini-2.5-Flash models.

\paragraph{Logicbench:}

For Logicbench, we used the first-order logic subset with a 160 total number of data samples. We added the following descriptions in the prompt placeholders (prompt templated shared in \ref{sec:datasettings}) `<special-instruction>' in prompt \#1:

\textit{ Question format should be: "context":<paragraph providing context for answering the question, pointing to reasoning for answering the question correctly compared to other incorrect choices>}

prompt \#2:

\textit{ Question format should be: "context":<paragraph providing context for answering the question, pointing to reasoning for answering the question correctly compared to other incorrect choices>}

The following demonstrates an example variant created for LogicBench.

\begin{quote}
\small
\textit{\textbf{Original question:} "context": If someone chooses to use public transportation, depending on their destination, they can ensure timely arrivals. taking the bus guarantees punctuality at the workplace, while opting for the train allows for attending college lectures. it should be noted that at least one of the following statements holds true: either (1) the individual named emily takes the bus to reach work or (2) she utilizes the train to commute to college.,
"question": Taking into account the context provided, what conclusion would be most appropriate?,
"choices": \{'choice\_1': 'They will choose to walk to their destination instead.', 'choice\_2': 'He will miss his train and be late for work.', 'choice\_3': 'she will get to the office on time. or she will attend her classes.', 'choice\_4': 'She will use a combination of both bus and train for her daily commute.'\}}

\textit{\textbf{SEM variant question:} "context": If someone chooses to use renewable energy sources, depending on their needs, they can ensure efficiency. Using solar panels guarantees a reduction in electricity bills at home, while opting for wind turbines allows for sustainable energy production on a farm. It should be noted that at least one of the following statements holds true: either (1) the individual named Alex uses solar panels to power his house or (2) he utilizes wind turbines to generate energy for his farm.,
"question": Considering the context provided, what conclusion would be most appropriate?,
"choices": \{'choice\_1': 'He will choose to rely entirely on fossil fuels instead.', 'choice\_2': 'He will face an unexpected power outage at home.', 'choice\_3': 'He will reduce his electricity bills. or he will produce sustainable energy.', 'choice\_4': 'He will use a combination of both solar panels and wind turbines for his energy needs.'\}}

\textit{\textbf{Right answer for both:} 'choice\_3'}

\end{quote}

In terms of evaluation prompts, we follow the prompt used in the source paper \cite{parmar-etal-2024-logicbench}.

\paragraph{BigBench-Hard:}

We used the temporal sequence subset from this dataset and the total number of data samples were 245. We added the following descriptions in the prompt placeholders (prompt template shared in \ref{sec:datasettings}) `<special-instruction>' in prompt \#1 and prompt \#2 respectively:

\textit{ Question format should be: "question":<question text comprising 5-7 sentences containing evidence needed to be all considered to arrive at the answer>}

\textit{ Question format should be: "question":<question text of similar length and complexity as the original question>
\\
"choices":<four choices>}

The following demonstrates an example variant created for BigBench-Hard.

\begin{quote}
\small
\textit{\textbf{Original question}: "question": Today, Emily went to the soccer field. Between what times could they have gone?
We know that:
Emily woke up at 10am.
Linda saw Emily waiting at the train station from 10am to 12pm.
Ashley saw Emily fixing their computer at the electronic store from 12pm to 1pm.
John saw Emily walking towards the Statue of Liberty from 1pm to 2pm.
The soccer field was closed after 8pm.
Between what times could Emily have gone to the soccer field?,
"choices": (A) 12pm to 1pm
(B) 10am to 12pm
(C) 2pm to 8pm
(D) 1pm to 2pm} 

\textit{\textbf{SEM variant question}: "question": A restaurant is open from 10am to 10pm and serves breakfast from 10am to 12pm, lunch from 12pm to 2pm, and dinner from 5pm to 10pm. The restaurant also has a happy hour from 2pm to 5pm. During the dinner hours, the restaurant offers a discount on all menu items. Considering the time intervals when the restaurant is not serving breakfast or lunch, which time frame does the dinner and happy hour fall into. The correct interval should include both happy hour and dinner service without overlapping with breakfast or lunch hours. Given these details about the operating hours of the restaurant, what time frame encompasses both happy hour and dinner service without including any part of breakfast or lunch service.,
"choices": (A) 12pm to 1pm
(B) 10am to 12pm
(C) 2pm to 8pm
(D) 1pm to 2pm} 

\textit{\textbf{Right answer for both:} '(C)'}
\end{quote}

In terms of evaluation prompts, we follow the prompt used in the source paper \cite{suzgun2022challenging}.

\paragraph{NLR-BIRD:} We filtered samples with table complexity >2 and <50 sampling 320 samples and passed them through the SEM pipeline. We added the following descriptions in the prompt placeholders (prompt templated shared in \ref{sec:datasettings}) `<special-instruction>' in prompt \#1:

\textit{ Question format should be: <question text for the given answer on the next line tabular data in format [[<column name 1>, <column name 2> .... ], [<row 1 cell 1 value>, <row1 cell 2 value>, ...], ...]containing the evidence for the answer>}

We added the following descriptions in the prompt placeholders `<special-instruction>' in prompt \#2:

\textit{ Question format should be: <question text of similar complexity as the original question and on the next line tabular data containing the same evidence for answering the question as the original question>}

The following demonstrates an example variant created for NLR-BIRD. NLR-BIRD contains a natural language (NL) question, tabular results for the question, and a response for that question based on the tabular results in NL - natural language response (NLR).

\begin{quote}
\small
\textit{\textbf{Original question:} "What is the category of event which was taken place in 2020-03-24T12:00:00?"}\\
\textit{\textbf{Tabular answer:} [['category'], ['Food'], ['Speaker Gifts'], ['Advertisement'], ['Parking']]}\\

\textit{\textbf{SEM variant question:} "What is the type of product that was launched on 2020-03-24T12:00:00?"}\\
\textit{\textbf{Tabular answer:} [['type'], ['Food'], ['Speaker Gifts'], ['Advertisement'], ['Parking']]}\\

\textit{\textbf{NLR for both:} 'Food, Speaker Gifts, Advertisement, Parking}'
\end{quote}

In terms of evaluation prompts and methods, we follow the details and combo-eval method used in the source paper \cite{singh-etal-2025-llms}.

\begin{table}[ht]
\centering
\begin{tabular}{llcc}
\hline
{Dataset} & Model & {Orig Acc} & {SEM Acc} \\
\hline
LogicBench & Ll4-Sct & 92.81& 92.26\\
LogicBench & Gm2.5fl & 85.00& 87.62 \\
\hline
BigBench-H & Ll4-Sct & 100.0 & 95.08\\
BigBench-H & Gm2.5fl & 99.19 & 94.23 \\
\hline
NLR-BIRD & Ll4-Sct & 95.86 & 98.91\\
NLR-BIRD & Gm2.5fl & 95.81 & 98.52 \\
\hline
\end{tabular}
\caption{Accuracy comparison between original and variant datasets}
\label{tab:dataset_variant_accuracy}
\end{table}

\section{GSM8K Benchmark Data Sample}
\label{sec:gsm8k_sample}

Two samples are shared for reference and context about the GSM8K data's structure.

Question:

\begin{quote}
\small
Benny saw a 10-foot shark with 2 6-inch remoras attached to it. 
What percentage of the shark's body length is the combined length of the remoras?
\end{quote}

Answer:
\begin{quote}
\small
First, find the combined length of the remoras in inches: 
6 inches/remora $\times$ 2 remoras = $\langle\langle 6 \times 2 = 12 \rangle\rangle$ 12 inches.

Then divide that number by 12 to convert it to feet: 12 inches / 12 inches/foot = <<1=1>>1 foot. 

Then divide the combined remora length in feet by the shark's length and multiply by 100\% to express the answer as a percentage: 1 foot / 10 feet * 100\% = 10\%. \#\#\#\# 10
\end{quote}

Question:
\begin{quote}
\small
Toula bakery and bought various types of pastries. She bought 3 dozen donuts which cost \$68 per dozen, 2 dozen mini cupcakes which cost \$80 per dozen, and 6 dozen mini cheesecakes for \$55 per dozen. How much was the total cost?
\end{quote}

Answer:
\begin{quote}
\small
The total charge for the doughnuts was 3 x \$68 = \$<<3*68=204>>204.

The total charge for the mini cupcakes was 2 x \$80 = \$<<2*80=160>>160.

The total charge for the mini cheesecakes was 6 x \$55 = \$<<6*55=330>>330.

Therefore the total amount Toula paid for the pastries was \$204 + \$160 + \$330 = \$<<204+160+330=694>>694.
\#\#\#\# 694
\end{quote}
\end{document}